%% file: main.tex
\documentclass[twocolumn]{svjour3}
\usepackage{subfiles}
\usepackage{soul}
\usepackage{times}
\usepackage{epsfig}
\usepackage{graphicx}
\usepackage{amsmath}
\usepackage{amssymb}
\usepackage{graphicx}
\usepackage{makecell}
\usepackage{rotating}
\usepackage{booktabs}
\usepackage{gensymb}
\usepackage[utf8]{inputenc} %
\usepackage[T1]{fontenc}    %
\usepackage{hyperref}       %
\usepackage{url}            %
\usepackage{booktabs}       %
\usepackage{amsfonts}       %
\usepackage{nicefrac}       %
\usepackage{microtype}      %
\usepackage{color}
\usepackage{subfigure}
\usepackage{makecell}
\usepackage{appendix}

\usepackage{pifont}
\newcommand{\cmark}{\color{green}\ding{51}}
\newcommand{\xmark}{\color{red}\ding{55}}

\usepackage{tabularx}
\usepackage{xspace}
\usepackage{balance}

\newcommand{\new}[1]{#1} %
\newcommand{\neww}[1]{#1} %
\newcommand{\newww}[1]{#1} %
\newcommand{\fin}[1]{#1} %
\newcommand{\prob}[1]{\textcolor{black}{#1}} %
\usepackage{xcolor}
\newcolumntype{C}[1]{>{\centering\let\newline\\\arraybackslash\hspace{0pt}}m{#1}}
\newcolumntype{Y}{>{\hsize=1.133\hsize\raggedright\arraybackslash}X} %
\newcolumntype{y}{>{\hsize=.6\hsize}X}

\newcommand{\dataSingleMultiFULL}{{\emph{Human        Optical Flow}}\xspace}
\newcommand{\dataSingleFULL}{     {\emph{Single-Human Optical Flow}}\xspace}
\newcommand{\dataMultiFULL}{      {\emph{Multi-Human  Optical Flow}}\xspace}
\newcommand{\dataSingle}{SHOF\xspace}
\newcommand{\dataMulti}{MHOF\xspace}

\newcommand{\citeMOSH}{{\cite{loper2014mosh,AMASS}\xspace}}
\newcommand{\dataSIGA}{{\emph{Embodied Hands}\xspace}}

\begin{document}

\title{Learning Multi-Human Optical Flow }

\author{
Anurag Ranjan                       \and
David T. Hoffmann                      \and
Dimitrios Tzionas                   \and
Siyu Tang                           \and
Javier Romero                       \and
Michael J. Black
\thanks{\textsuperscript{*}Anurag Ranjan and David Hoffmann contributed equally.}
}

\institute{Anurag Ranjan\textsuperscript{*}$^1$ \at
        {anurag.ranjan@tue.mpg.de}           %
           \and
        David T. Hoffmann\textsuperscript{*}$^1$ \at
        {david.hoffmann@tue.mpg.de}
            \and
        Dimitrios Tzionas$^1$ \at
        {dimitris.tzionas@tue.mpg.de}
            \and
        Siyu Tang$^1$ \at
        {siyu.tang@tue.mpg.de}
            \and
        Javier Romero$^2$ \at
        {javier@amazon.com}
            \and
        Michael J. Black$^1$ \at
        {black@tue.mpg.de}
        \and
        $^1$ Max Planck Institute for Intelligent Systems, Germany \\
        $^2$ Amazon Inc.
        \\
        {\scriptsize
        This work was done when JR was at MPI-IS. MJB has received research gift funds from Intel, Nvidia, Adobe, Facebook, and Amazon. While MJB is a part-time employee of Amazon, his research was performed solely at, and funded solely by, MPI.
        \neww{MJB has financial interests in Amazon and Meshcapade GmbH.}
        }
}

\date{Received: date / Accepted: date}

\maketitle

\begin{abstract}
\subfile{abstract}
\end{abstract}

\section{Introduction}
\subfile{1intro}

\section{Related Work}
\subfile{2related_works}

\section{The Human Optical Flow Dataset}
\subfile{3human_flow_dataset_final}

\vspace{2em}
\section{Learning}
\subfile{4learning}

\section{Experiments}
\subfile{5experiments}

\section{Conclusion and Future Work}
\subfile{6conclusion_javier}

\section*{Acknowledgements}
We thank Yiyi Liao for helping us with optical flow evaluation. 
\newww{
We thank Sergi Pujades for helping us with collision detection of meshes. 
}
We thank Cristian Sminchisescu for the Human3.6M MoCap marker data. 

\balance
\bibliographystyle{unsrt}
\bibliography{egbib}
\end{document}

%% file: abstract.tex
The optical flow of humans is well known to be useful for the analysis of human action. \neww{Recent optical flow methods focus on training deep networks to approach the problem. However, the training data used by them does not cover the domain of human motion. 
Therefore, we develop a dataset of multi-human optical flow and train optical flow networks on this dataset.}
We use a 3D model of the human body and motion capture data to synthesize realistic flow fields \new{in both single- and multi-person images}.
We then train optical flow networks to estimate human flow fields from pairs of images.
We demonstrate that our trained networks \new{are} more accurate than a wide range of \prob{top methods} on held-out test data 
and that \new{they can} generalize well to real image sequences.
The code, trained \new{models} and the dataset are available for research.

%% file: 1intro.tex
\begin{figure*}[t]
\centerline{
\includegraphics[width=\linewidth]{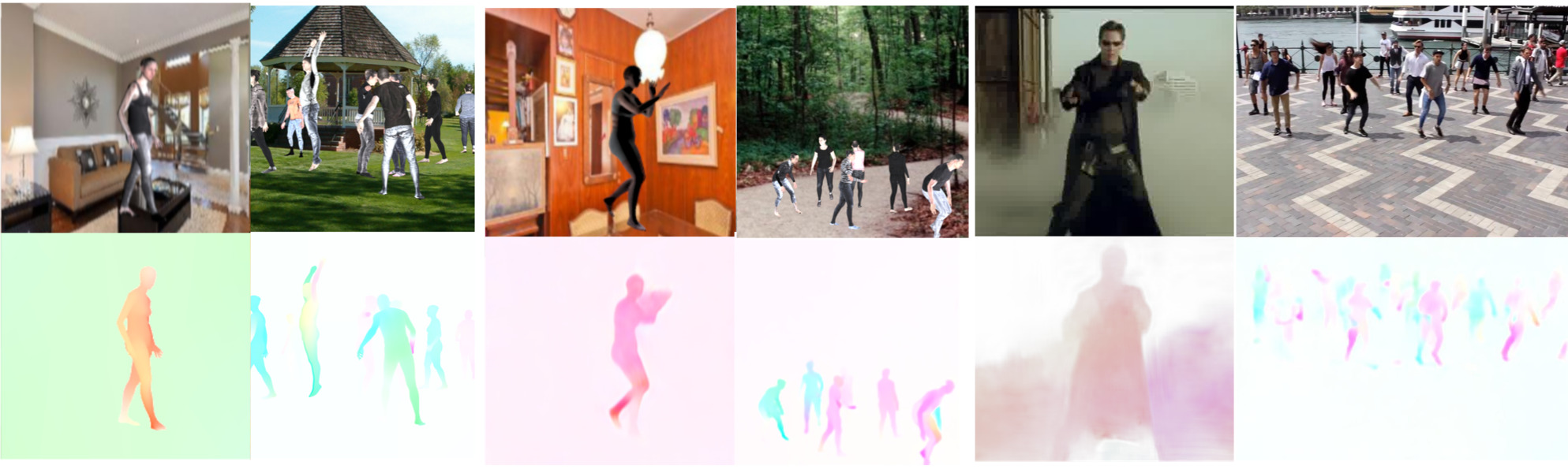}
}
\begin{flushleft}
\hspace{0.7in} \small{(a) Our dataset \hspace{0.9in} (b) Results on synthetic scenes \hspace{0.8in} (c) Results on real world scenes}
\end{flushleft}
\caption{(a) We simulate human motion in virtual worlds creating an extensive dataset with images (top row) and flow fields (bottom row); color coding from \cite{baker2011database}. (b) We train SPyNet \cite{ranjan2016optical} and PWC-Net \cite{Sun2018PWC-Net} for human motion estimation  and show that they \fin{perform} better when trained on our dataset and (c) can generalize to human motions in real world scenes. Columns show single-person and multi-person cases alternately.}
\label{fig:teaser}
\end{figure*}

A significant fraction of videos on the Internet contain people moving \cite{scienceselfies} and the literature suggests that optical flow plays an important role in understanding human action \cite{Jhuang:ICCV:2013,soomro2012ucf101}.
Several action recognition datasets \cite{soomro2012ucf101,kuehne2011hmdb} contain human motion as a major component.
The 2D motion of humans in video, or {\em human optical flow}, is an important feature that provides a building block for
systems that can understand and interact with humans.
Human optical flow is useful for various applications including analyzing pedestrians in road sequences, motion-controlled gaming, activity recognition, human pose estimation,  etc.

Despite this, optical flow has previously been treated as a generic, low-level, vision problem.
Given the importance of people, and the value of optical flow in understanding them, \neww{we develop a dataset and trained models that are specifically tailored to humans and their motion}.
Such motions are non-trivial since humans are complex, articulated objects that vary in shape, size and appearance.
They move quickly, adopt a wide range of poses, and self-occlude \new{or occlude in multi-person scenarios}.

Our goal is to obtain more accurate 2D motion estimates for human bodies by training a flow algorithm specifically for human movement.
To do so, we create a large and realistic dataset of humans moving in virtual worlds with ground truth optical flow (Fig.~\ref{fig:teaser}(a)), \neww{called the \dataSingleMultiFULL dataset}.
\new{This is comprised of two parts; the \dataSingleFULL dataset (\dataSingle),
where the image sequences contain only one person in motion and the \dataMultiFULL dataset (\dataMulti)
where images contain multiple people involving significant occlusion between them.}
\neww{We analyse the performance of SPyNet \cite{ranjan2016optical} and PWC-Net \cite{Sun2018PWC-Net} by training (fine-tuning) them on both the \dataSingle and \dataMulti dataset. We observe that the optical flow performance of the networks improves on sequences containing human scenes, both qualitatively and quantitatively.}
Furthermore we show that the trained networks generalize to real video sequences  (Fig.~\ref{fig:teaser}(c)).
Several datasets and benchmarks \cite{baker2011database,Geiger2012CVPR,Butler:ECCV:2012} have been established to drive the progress in optical flow.
We argue that these datasets are insufficient for the task of human motion estimation and, despite its importance, no attention has been paid to datasets and \neww{models} for human optical flow.
One of the main reasons is that dense human motion is extremely difficult to capture accurately in real scenes.
Without ground truth, there has been little work focused specifically on estimating human optical flow.
To advance research on this problem, the community needs a dataset tailored to human optical flow.

A key observation is that recent work has shown that optical flow methods trained on synthetic data
\cite{ranjan2016optical,dosovitskiy2015flownet,ilg2016flownet} generalize relatively well to real data.
Additionally, these methods obtain state-of-the-art results with increased realism of the training data \cite{flyingthings,Gaidon:Virtual:CVPR2016}.
This motivates our effort to create a dataset designed for human motion.

To that end, we use \neww{the SMPL \cite{Bogo:ECCV:2016} and SMPL+H \cite{MANO:SIGGRAPHASIA:2017} models, that capture the human body alone and the body together with articulated hands respectively,} to generate different human shapes \neww{including hand and finger motion}.
We then place \neww{humans} on random indoor backgrounds and simulate human activities like running, walking, dancing etc. using motion capture data \citeMOSH.
Thus, we create a large virtual dataset that captures the statistics of natural human motion \new{in multi-person scenarios}.
We then train optical flow networks \neww{on this} dataset and evaluate their performance for estimating human motion.
While the dataset can be used to train any flow method, we \neww{focus specifically on networks based on spatial pyramids, namely SpyNet \cite{ranjan2016optical} and PWC-Net \cite{Sun2018PWC-Net},} because they are compact and computationally efficient.

\new{A preliminary version of this work appeared in \cite{Ranjan:BMVC:2018} that presented a dataset and model for human optical flow for the \emph{single-person} case with a \emph{body-only} model.
The present work extends \cite{Ranjan:BMVC:2018} for the \emph{multi-person} case, as images with multiple occluding people have different statistics.
It further employs a holistic model of the \emph{body together with hands} for more realistic motion variation.
This work also extends training SPyNet \neww{\cite{ranjan2016optical}} and PWC-Net \neww{\cite{Sun2018PWC-Net}} using the new dataset in contrast to training only SPyNet in the earlier work \cite{Ranjan:BMVC:2018}.
Our experiments show both qualitative and quantitative improvements.
}

In summary, our major contributions in this extended work are:
1) We provide the \dataSingleFULL dataset (\dataSingle) of human bodies in motion with realistic textures and backgrounds, having  $146,020$ frame pairs \new{for single-person scenarios}.
2) \new{We provide} {the} \new{\dataMultiFULL dataset (\dataMulti), with $111,312$ frame pairs of multiple human bodies in motion, with improved textures and realistic visual occlusions, but without (self-)collisions or intersections of body meshes.}
These two datasets together comprise the \dataSingleMultiFULL dataset.
3) \neww{We fine-tune SPyNet \cite{Ranjan:BMVC:2018} on \dataSingle and show that its performance improves by about $43\%$ (over the initial SPyNet), while it also outperforms existing state of the art by about $30\%$.}
\neww{Furthermore, we fine-tune SPyNet and PWC-Net on \dataMulti and observe improvements of $10-20\%$ (over the initial SPyNet and PWC-Net).}
\neww{Compared to existing state of the art, improvements are particularly high for human regions. After masking out the background, we observe improvements of up to $13\%$ for human pixels.}
4) We provide the dataset \neww{files, dataset rendering} code, \neww{training code} and trained models\footnote{\url{https://humanflow.is.tue.mpg.de}} for research purposes.

%% file: 2related_works.tex
\textbf{Human Motion.} %
Human motion can be understood from 2D motion. 
Early work focused on the movement of 2D joint locations \cite{Johansson1973}  or simple motion history images \cite{mhi_davis_2001}.
Optical flow is also a useful cue.
Black et al.~\cite{Black:IEEE:1997} use principal component analysis (PCA) to parametrize human motion but use noisy flow computed from image sequences for training data.
More similar to us, Fablet and Black \cite{Black:ECCV:2002} use a 3D articulated body model and motion capture data to project 3D body motion into 2D optical flow.
They then learn a view-based PCA model of the flow fields.  
We use a more realistic body model to generate a large dataset and use this to train a CNN to directly estimate dense human flow from images.

Only a few works in pose estimation have exploited human motion and, in particular\neww{,} several methods \cite{Fragkiadaki2013,Zuffi:ICCV:2013} use optical flow constraints to improve 2D human pose estimation in videos.
Similar work~\cite{PfisterCZ15,CharlesPMHZ16} propagates pose results temporally using optical flow to encourage time consistency of the estimated bodies. 
Apart from its application in warping between frames, the structural information existing in optical flow \neww{alone} has been used for pose estimation %
\cite{flowcap} or in conjunction with an image stream ~\cite{FeichtenhoferPZ16,Dong_2018_CVPR}.

\textbf{Learning Optical Flow.} 
There is a long history of optical flow estimation, which we do not review here.
Instead, we focus on the relatively recent literature on learning flow.
Early work looked at learning flow using Markov Random Fields \cite{Freeman2000}, PCA \cite{wulff2015efficient}, or shallow convolutional models \cite{roth2008learning}.
Other methods also combine learning with traditional approaches, formulating flow as a discrete \cite{guney2016ACCV} or continuous \cite{epicflow} optimization problem. 

The most recent methods employ large datasets to estimate optical flow using deep neural networks. Voxel2Voxel \cite{Tran:End2End:2016} is based on volumetric convolutions to predict optical flow using $16$ frames simultaneously but does not perform well on benchmarks.
Other methods \cite{ranjan2016optical,dosovitskiy2015flownet,ilg2016flownet} compute two frame optical flow using an end-to-end deep learning approach. FlowNet \cite{dosovitskiy2015flownet} uses the Flying Chairs dataset \cite{dosovitskiy2015flownet} to compute optical flow in an end-to-end deep network. 
FlowNet $2.0$ \cite{ilg2016flownet} uses stacks of networks from FlowNet and performs significantly better, particularly for small motions.
Ranjan and Black \cite{ranjan2016optical} propose a Spatial Pyramid Network that employs a small neural network on each level of an image pyramid to compute optical flow. Their method uses a much smaller number of parameters and achieves similar performance as FlowNet \cite{dosovitskiy2015flownet} using the same training data. Sun et al. \cite{Sun2018PWC-Net} use image features in a similar spatial pyramid network achieving state-of-the-art results on optical flow benchmarks.
Since the above methods are not trained with human motions, they do not perform well on our Human Optical Flow dataset.

\textbf{Optical Flow Datasets.} 
Several datasets have been developed to facilitate training and benchmarking of optical flow methods.
Middlebury is limited to small motions \cite{baker2011database}, KITTI is focused on rigid scenes and automotive motions \cite{Geiger2012CVPR}, while Sintel has a limited number of synthetic scenes  \cite{Butler:ECCV:2012}.
These datasets are mainly used for evaluation of optical flow methods and are generally too small to support training neural networks.

To learn optical flow using neural networks, more datasets have emerged that contain examples on the order of tens of thousands of frames. 
The Flying Chairs \cite{dosovitskiy2015flownet} dataset contains about $22,000$ samples of chairs moving against random backgrounds. 
Although it is not very realistic or diverse, it provides training data for neural networks \cite{ranjan2016optical,dosovitskiy2015flownet} that achieve reasonable results on optical flow benchmarks. 
Even more recent datasets \cite{flyingthings,Gaidon:Virtual:CVPR2016} for optical flow are especially designed for training deep neural networks. 
Flying Things \cite{flyingthings} contains tens of thousands of samples of random 3D objects in motion. 
\fin{The Creative Flow+ Dataset \cite{shugrina2019creative} contains diverse artistic videos in multiple styles.}
The Monkaa and Driving scene datasets \cite{flyingthings} contain frames from animated scenes and virtual driving respectively. 
Virtual KITTI \cite{Gaidon:Virtual:CVPR2016} uses graphics to generate scenes like those in KITTI and is two orders of magnitude larger. 
Recent synthetic datasets \cite{Gaidon2014} show that synthetic data can train networks that generalize to real scenes.

For human bodies, 
\fin{some works \cite{barbosa2018looking,ghezelghieh2016learning} render images with the non-learned artist-defined MakeHuman model \cite{makehuman} for 3D pose estimation or person re-identification, correspondingly.}
\fin{However, statistical parametric models learned from 3D scans of a big human population, like SMPL \cite{SMPL:2015}, capture the real distribution of human body shape.} 
The SURREAL dataset \cite{surreal} uses 3D {SMPL} human meshes rendered on top of \new{color} images to train networks for depth estimation, and body part segmentation. 
While not fully realistic, they show that this data is sufficient to train methods that generalize to real data.
We go beyond \new{these} works to address the problem of optical flow.

%% file: 3human_flow_dataset_final.tex
\begin{figure*}
\centerline{
\includegraphics[width=0.9\textwidth]{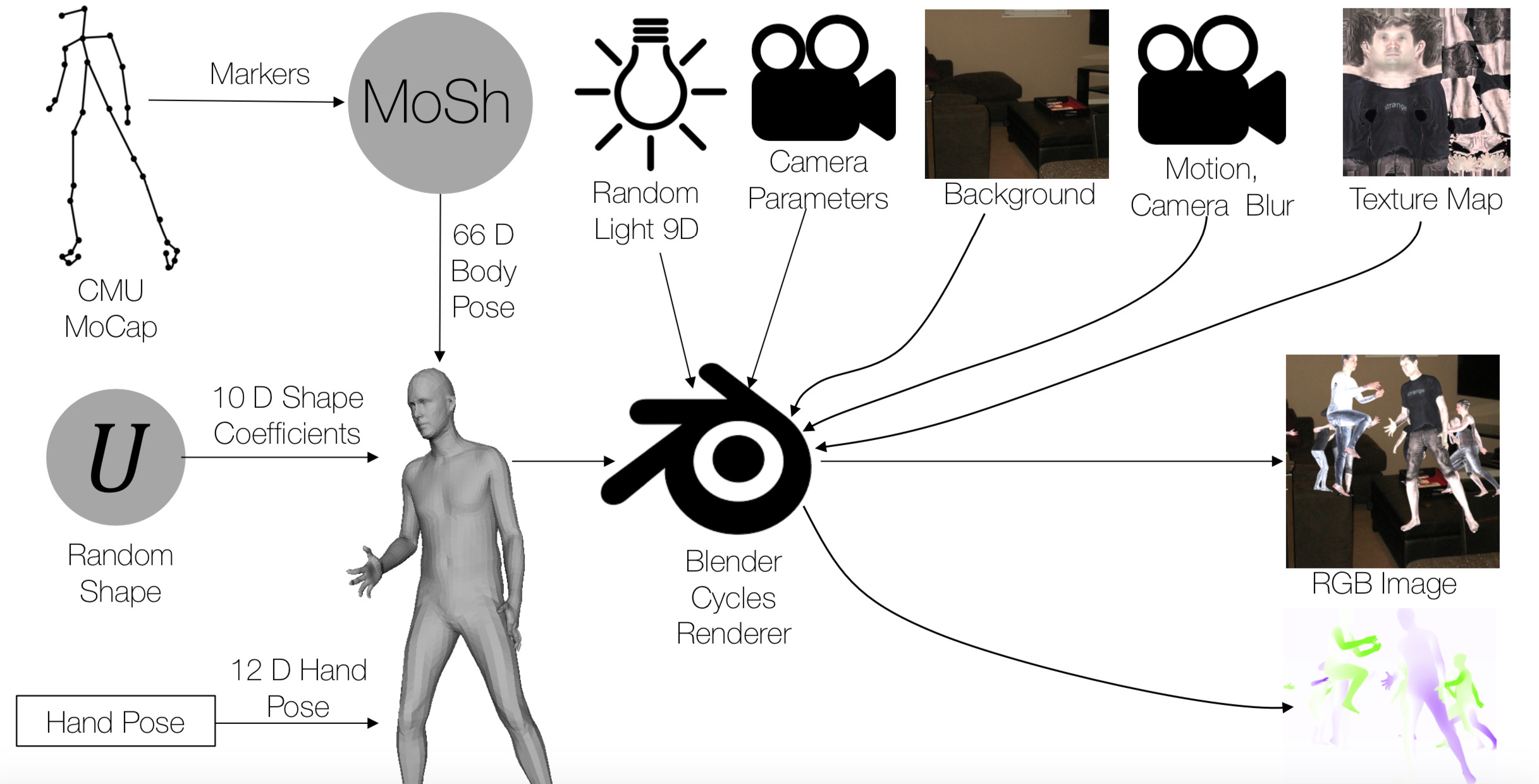}
}
\vspace{-0.2cm}
\caption{Pipeline for generating the RGB frames and ground truth optical flow for the \emph{Multi-Human Optical Flow dataset}. 
\neww{The datasets used in this pipeline are listed in Table \ref{tab:resources},} 
\neww{while the various rendering component are summarized in Table \ref{tab:methods}}. 
}
\label{fig:pipe}
\vspace{-0.4cm}
\end{figure*}

Our approach generates a realistic dataset of synthetic human motions by simulating them against different realistic backgrounds. 
We use \new{parametric models \cite{MANO:SIGGRAPHASIA:2017,SMPL:2015} to generate \new{synthetic humans} with} a wide variety of different human shapes. %
\neww{We employ Blender\footnote{\url{https://www.blender.org}} and its Cycles rendering engine to generate realistic synthetic image frames and optical flow. 
In this way we create the \dataSingleMultiFULL  dataset, that 
}
\new{is comprised of two parts.}
We first create the \emph{Single-Human Optical Flow} (\dataSingle) dataset \cite{Ranjan:BMVC:2018} using the body-only SMPL model \cite{SMPL:2015} in images containing a single synthetic human. 
However, image statistics are different for the single- and multi-person case, as multiple people tend to occlude each other in complicated ways.
For this reason we then create the \emph{Multi-Human Optical Flow} (\dataMulti) dataset to better capture this realistic interaction. 
To make images even more realistic \neww{for \dataMulti}, we replace SMPL \cite{SMPL:2015} with the SMPL+H \cite{MANO:SIGGRAPHASIA:2017} model
that models the body together with \neww{articulated} fingers, to have richer motion variation.
\new{In the rest of this section, we describe the components of our rendering pipeline, shown in Figure \ref{fig:pipe}.}
\neww{For easy reference, 
in Table \ref{tab:resources} we summarize the data used to generate the \dataSingle and \dataMulti datasets, while in Table ~\ref{tab:methods} we summarize the various tools, Blender passes and parameters used for rendering. In the rest of the section, we describe the modules used for generating the data.}

\textbf{\subsection{Human Body Generation}}
\textbf{Body Model.} %
\neww{A parametrized body model is necessary to generate human bodies in a scene.}
\neww{In the \dataSingle dataset, we use} SMPL \cite{SMPL:2015} \neww{for generating human body shapes}. \neww{For the \dataMulti dataset we, use} SMPL+H \cite{MANO:SIGGRAPHASIA:2017} \neww{that parametrizes the human body together with articulated fingers, for increased realism.} 
\neww{The models} 
are parameterized by pose and shape parameters to change the body posture and identity, as shown in Figure \ref{fig:pipe}.
They also contain a UV appearance map that allows us to change the skin tone, face features and clothing texture of the \new{resulting virtual humans}.\xspace%

\textbf{Body Poses.}
\neww{The next step is articulating the human body with different poses, to create moving sequences.
To find such poses, we use 3D MoCap datasets \cite{h36m_pami,cmumocapdatabase,Sigal:IJCV:10b} 
that capture 3D MoCap marker positions, glued onto the skin surface of real human subjects. 
We then employ MoSh \citeMOSH~ that fits our body model to these 3D markers by optimizing over parameters of the body model for articulated pose, translation and shape. 
The pose specifically is a vector of axis-angle parameters, that describes how to rotate each body part around its corresponding skeleton joint. 
}

\new{For the \dataSingle dataset, we use} the Human3.6M dataset \cite{h36m_pami}. %
that contains five subjects for training (S1, S5, S6, S7, S8) and two for testing (S9, S11).
Each subject performs $15$ actions twice, resulting in $1,559,985$ frames for training and $550,727$ for testing.
These sequences are subsampled at a rate of $16\times$, resulting in $97,499$ training and $34,420$ testing poses from Human3.6M.

\new{For the \dataMulti dataset, we use the CMU \cite{cmumocapdatabase} and HumanEva \cite{Sigal:IJCV:10b} MoCap datasets to increase motion variation.}
\new{From CMU MoCap dataset, we use $2,605$ sequences of $23$ high-level action categories. 
From the HumanEva dataset, we use more than $10$ sequences performing actions from 6 different action categories.} %
\new{To reduce redundant poses \neww{and allow for larger motions between frames}, sequences are subsampled to $12$ fps resulting in \new{$321,873$ poses}. 
As a result the final \dataMulti dataset has $254,211$ poses for training, $32,670$ for validation and $34,992$ for testing.
}

\textbf{Hand Poses.}
\new{
Traditionally MoCap systems and datasets \cite{h36m_pami,cmumocapdatabase,Sigal:IJCV:10b} record the motion of body joints, %
and avoid the tedious capture of detailed hand and finger motion. 
However, in natural settings, people use their body, hands and fingers to communicate social cues and to interact with the physical world. 
\neww{To enable our methods to learn such subtle motions, it should be represented in our training data.}
Therefore, we 
use the SMPL+H model \cite{MANO:SIGGRAPHASIA:2017} and 
augment the body-only MoCap datasets, described above, with finger motion. Instead of using random finger poses that would generate unrealistic optical flow, we employ the \dataSIGA~dataset \cite{MANO:SIGGRAPHASIA:2017} and sample continuous finger motion to generate realistic optical flow. 
We use $43$ sequences of hand motion with $37,232$ frames recorded at $60$ Hz by \cite{MANO:SIGGRAPHASIA:2017}.
Similarly to body MoCap, we subsample hand MoCap to $12$ fps to reduce overlapping poses without sacrificing variability.
}

\textbf{Body Shapes.}
\neww{Human bodies vary a lot in their proportions, since each person has a unique body shape. 
To represent this in our dataset, we first learn a gender specific Gaussian distribution of shape parameters, by fitting SMPL to 3D CAESAR scans \cite{robinette2002civilian} of both genders. 
We then sample random body shapes from this distribution to generate a large number of realistic body shapes for rendering. 
However, naive sampling can result in extreme and unrealistic shape parameters, therefore we bound the shape distribution to avoid unlikely shapes.} 

For the \dataSingle dataset we bound the shape parameters to the range of $[-3,3]$ standard deviations for each shape coefficient and draw a new shape for every subsequence of $20$ frames to increase variance.

\neww{
For the \dataMulti dataset, \neww{we account explicitly for collisions and intersections}, since intersecting virtual humans would result in generation of inaccurate optical flow. 
To minimize such cases, we use similar sampling as above with only small differences. 
We first use shorter subsequences of $10$ frames for less frequent inter-human intersections. 
Furthermore, we bound the shape distribution to the narrower range of \new{$[-2.7,2.7]$} standard deviations, since re-targeting motion to unlikely body shapes is more prone to mesh self-intersections. }

\begin{table*}[t]
\setlength\extrarowheight{2.0pt}
    \centering
    \begin{tabularx}{\textwidth}{yYYY}
        
                    \toprule
                     & \dataSingle & \dataMulti & Purpose\\
                    \midrule
                    MoCap data & Human3.6M \cite{h36m_pami} & CMU \cite{cmumocap}, HumanEva \cite{Sigal:IJCV:10b} & Natural body poses \\
                    
                    MoCap $\rightarrow$ SMPL& MoSh \citeMOSH & MoSh \citeMOSH & SMPL parameters from MoCap\\
                  
                    Training poses                       & $97,499$ & $254,211$ & Articulate virtual humans \\
                  
                    Validation poses                       & -- & $32,670$ & Articulate virtual humans \\
                  
                    Test poses                       & $34,420$ & $34,992$ & Articulate virtual humans \\
                  
                    Hand pose dataset & -- & Embodied Hands \cite{MANO:SIGGRAPHASIA:2017} & Natural finger poses \\
                  
                    Body shapes  & Sample Gaussian distr. (CAESAR) bounded within $[-3,3]$ st.dev. &                        Sample Gaussian distr. (CAESAR) bounded within $[-2.7,2.7]$ st.dev. & Body proportions of virtual humans \\ 
                  
                    Textures & CAESAR, \newline non-CAESAR & CAESAR (hands improved), \newline non-CAESAR (hands improved) & Appearance of virtual humans\\
                   
                    Background & LSUN~\cite{yu_lsun} (indoor) \newline $417,597$ images & SUN397~\cite{xiao2010sun} (indoor and outdoor) \newline $30,022$ images & Scene background \\
                    \bottomrule
    \end{tabularx}
    \caption{\neww{Comparison of datasets and most important data preprocessing steps used to generate the \dataSingle and \dataMulti datasets. A short description of the respective part is provided in the last column.}}
    \label{tab:resources}
\end{table*}

\begin{table*}[t]
\setlength\extrarowheight{2pt}
    \centering
    \begin{tabularx}{\textwidth}{yYYY}
                
                    \toprule
                     & \dataSingle & \dataMulti & Purpose\\
                    \midrule

                    Rendering & Cycles & Cycles & Synthetic RGB image rendering \\
                  
                    Optical flow & Vector pass (Blender) & Vector pass (Blender) & Optical flow ground truth\\                  
 
                    Segment. masks & Material pass (Blender) & Material pass (Blender) & Body part segment. masks (Fig.~\ref{fig:segmentation})\\
                    
                    Motion blur & Vector pass (Blender) & Vector pass (Blender) & Realistic motion blur artifacts \\
                  
                    Imaging noise & Gaussian blur (pixel space) \newline $1$px std.dev. for $30$\% of images & Gaussian blur (pixel space) \newline $1$px std.dev. for $30$\% of images & Realistic image imperfections\\

                   Camera translation & Sampled for 30\% of frames from Gaussian with 1 cm {std.dev.} & Sampled for 30\% of subsequences from Gaussian with 1 cm {std.dev.} & Realistic perturbations of the camera (and resulting optical flow)\\
                   
                   Camera rotation & Sampled per frame from Gaussian with 0.2 degrees {std.dev.} & -- & Realistic perturbations of the camera (and resulting optical flow) \\
                  
                    Illumination & Spherical harmonics \cite{spherical_harmonics} & Spherical harmonics \cite{spherical_harmonics} & Realistic lighting model\\
                   
                    Subsequence length & $20$ frames & $10$ frames & Number of successive frames with consistent rendering parameters \\
                    
                    Mesh collision & -- &  BVH \cite{collisionDeformableObjects} & Detect (self-)collisions on the triangle level to avoid defect Optical Flow\\
                    
                    \bottomrule
    \end{tabularx}
    \caption{
        \neww{Comparison of tools, Blender passes and parameters used to generate the \dataSingle and \dataMulti datasets. The last column provides a short description of the respective method. 
        }
    }
    \label{tab:methods}
\end{table*}

\textbf{Body Texture.}
\neww{We use
the CAESAR dataset \cite{robinette2002civilian} to generate a variety of human skin textures}.
Given SMPL registrations to CAESAR scans, the original per-vertex color in the CAESAR dataset is transferred into the SMPL texture map. 
Since fiducial markers were placed on the bodies of CAESAR subjects, we remove them from the textures and inpaint them to produce a natural texture. %
\new{In total, we use 166 CAESAR textures that are of good quality.}
The main drawback of CAESAR scans is their homogeneity in terms of outfit, since all of the subjects wore grey shorts and the women wore sports bras. In order to increase the clothing variety, we also use textures extracted from our 3D scans (referred as \new{non-CAESAR in the following), to which we register SMPL with 4Cap \cite{Dyna:SIGGRAPH:2015}}.
A total of $772$ textures from $7$ different subjects with different clothes were captured. 
We anonymized the textures by replacing the face by the average face in CAESAR, after correcting it to match the skin tone of the texture.
\new{Textures are grouped according to the gender, which is randomly selected for each virtual human.}

For the \dataSingle dataset the textures were split in training and testing sets with a $70/30$ ratio, while each texture dataset is sampled with a $50\%$ chance. 
\new{For the \dataMulti dataset, we introduce more refined splitting with a $80/10/10$ ratio for the train, validation and test sets. Moreover, since we introduce also finger motion, we want to favour sampling non-CAESAR textures, due to the bad quality of CAESAR texture maps for the finger region.
Thus each texture is sampled with equal probability.}

\new{
\textbf{Hand Texture.}
Hands and fingers are hard to be scanned due to occlusions and measurement limitations.
As a result, texture maps are particularly noisy or might even have holes.
Since texture is important for optical flow, we augment the body texture maps to improve hand regions.
For this we follow a divide and conquer approach. First, we capture hand-only scans with a 3dMD scanner \cite{MANO:SIGGRAPHASIA:2017}. Then, we create hand-only textures using the MANO model \cite{MANO:SIGGRAPHASIA:2017}, getting $176$ high resolution textures from $20$ subjects. Finally, we use the hand-only textures to replace the problematic hand regions in the full-body texture maps. 

We also need to find the best matching hand-only texture for every body texture. Therefore, we convert all texture maps in HSV space, and compute the mean HSV value for each texture map from standard sampling regions.
For full body textures, we sample face regions without facial hair;
while for hand-only textures, we sample the center of the outer palm. 
Then, for each body texture map we find the closest hand-only texture map in HSV space, and shift the values of the latter by the HSV difference, so that the hand skin tone becomes more similar to the facial skin tone. 
Finally, this improved hand-only texture map is used to replace the pixels in the hand-region of the full body texture map. 
}

\textbf{(Self-) Collision.}
\new{
The \dataMulti dataset contains multiple virtual humans moving differently, so there are high chances of collisions and penetrations. 
This is undesirable because penetrations are physically implausible and unrealistic. Moreover, the generated ground truth optical flow might have artifacts. 
Therefore, we employ a collision detection method to avoid intersections and penetrations. %

Instead of using simple bounding boxes for rough collision detection, we draw inspiration from \cite{Tzionas:IJCV:2016} and perform accurate and efficient collision detection on the triangle level using bounding volume hierarchies (BVH) \cite{collisionDeformableObjects}.
This level of detailed detection allows for challenging occlusions with small distances between virtual humans, that can commonly be observed for realistic interactions between real humans.
This method is useful not only for inter-person collision detection, but also for self-intersections. 
This is especially useful for our scenarios, as re-targeting body and hand motion to people of different shapes might result in unrealistic self-penetrations. 
The method is applicable out of the box, with the only exception that we exclude checks of neighboring body parts that are always or frequently in contact, e.g. upper and lower arm, or the two thighs. 
}

\textbf{\subsection{Scene Generation}
}
\textbf{Background texture.}
For the scene background in the \dataSingle dataset, we use random indoor images from the LSUN dataset~\cite{yu_lsun}. This provides a good compromise between simplicity and the complex task of generating varied full 3D environments.
We use $417,597$ images from the LSUN categories kitchen, living room, bedroom and dining room.
These images are placed as billboards, $9$ meters from the camera, and are not affected by the spherical harmonics lighting.

\new{In the \dataMulti dataset, we increase the variability in background appearance,  
We employ %
the Sun397 dataset \cite{xiao2010sun} that contains images for $397$ highly variable scenes that \neww{are both} indoor and outdoor, \neww{in contrast to LSUN}. 
For quality reasons, we reject all images with resolution smaller than $512 \times 512$ px, and also reject images that contain humans using mask-RCNN \cite{he2017mask,matterport_maskrcnn_2017}. %
As a result, we use $30,222$ images, split in $24,178$ for the training set and $3,022$ for each of the validation and test sets.}
\new{Further, we increase the distance between the camera and background to $12$ meters, to increase the space in which \neww{the multiple} virtual humans can move without colliding frequently with each other, while still being close enough for visual occlusions.}

\textbf{Scene Illumination.}
We illuminate the bodies with Spherical Harmonics lighting~\cite{spherical_harmonics} that define basis vectors for light directions.
This parameterization is useful for randomizing the scene light by randomly sampling the coefficients with a bias towards natural illumination. 
The coefficients are uniformly sampled between $-0.7$ and $0.7$, apart from the ambient illumination, 
which has a minimum value of $0.3$ to avoid extremely dark images, and 
illumination direction, 
which is strictly negative 
\neww{to favour illumination coming from above}.

\textbf{Increasing Image Realism.}
In order to increase realism, we introduced three types of image imperfections. 
First, for $30\%$ of the generated images we introduced camera motion between frames. 
This motion perturbs the location of the camera with Gaussian noise of $1$ cm standard deviation between frames and rotation noise of $0.2$ degrees standard deviation per dimension in an Euler angle representation.
Second, we add motion blur to the scene using the {Vector Blur Node} in Blender, and integrated over $2$ frames sampled with $64$ steps between the beginning and end point of the motion.
Finally, we add a Gaussian blur to $30\%$ of the images with a standard deviation of $1$ pixel. 
\new{%
}

\textbf{Scene Compositing.}
\new{For animating virtual humans, each MoCap sequence is selected at least once. 
To increase variability, each sequence is split into subsequences. 
For the first frame of each subsequence, we sample a body and background texture, lights, blurring and camera motion parameters, and re-position virtual humans on the horizontal plane. We then 
introduce a random rotation around the $z$-axis for variability in the motion direction. 

For the \dataSingle dataset, we use subsequences of $20$ frames, and at the beginning of each one the single virtual human is re-positioned in the scene such that the pelvis is projected onto the image center. }
\new{%

For the \dataMulti dataset, we increase the variability with smaller subsequences of $10$ frames and introduce more challenging visual occlusions by uniformly sampling the number of virtual humans in the range $[4,8]$.
We sample MoCap sequences $S_j$ with a probability of $p_j=\frac{|S_j|}{\sum_{i=1}^{|S|}|S_i|}$, where $|S_j|$ denotes the number of frames of sequence $S_j$ and $|S|$ the number of sequences. 
}
\new{In contrast to the \dataSingle dataset, for the \dataMulti dataset the virtual humans are not re-positioned at the center, as they would all collide. 
Instead, they are placed at random locations on the horizontal plane within camera visibility, making sure there are no collisions with other virtual humans or the background plane during the whole subsequence. 
}

\textbf{\subsection{Ground Truth Generation}}

\begin{figure}
\centering     %
\label{fig:segmentation}\includegraphics[width=0.42\textwidth]{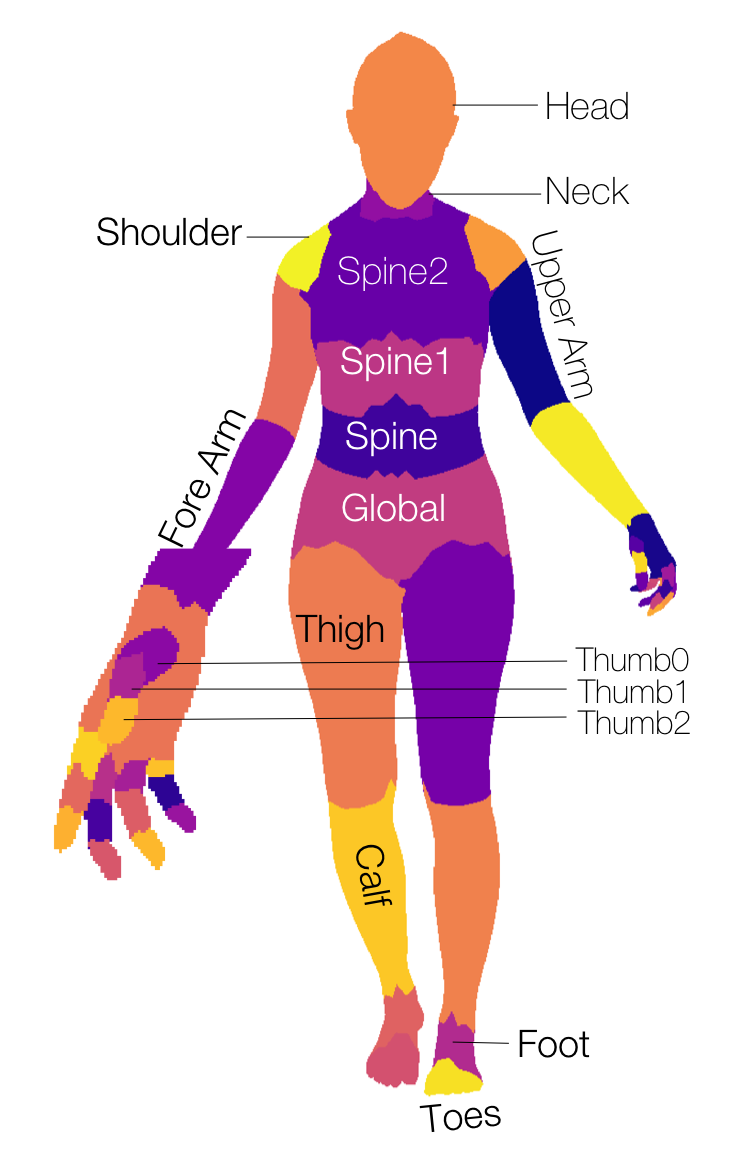}
\caption{Body part segmentation for the \new{SMPL+H} model. Symmetrical body parts are labeled only once. Finger joints follow the same naming convention as shown for the thumb. (Best viewed in color)}
\end{figure}

\textbf{Segmentation Masks.}
\new{%
Using the material pass of Blender, we store for each frame the ground truth body part segmentation for our models. %
Although the body part segmentation for both models is similar, SMPL models the palm and fingers as one part, while SMPL+H has a different part segment for each finger bone. 
Figure \ref{fig:segmentation} shows an example body part segmentation for SMPL+H. 
These segmentation masks allow us to perform a per body-part evaluation of our optical flow estimation. 
}

\neww{
\textbf{Rendering \& Ground Truth Optical Flow}. 
For generating images, we use the open source suite Blender and its \emph{vector pass}.
The render pass is typically used for producing motion blur, and it produces the motion in image space of every pixel; i.e. the ground truth optical flow. We are mainly interested in the result of this pass, together with the color rendering of the textured bodies.
}

%% file: 4learning.tex
\begin{figure*}[t]
\centerline{
\includegraphics[width=\linewidth]{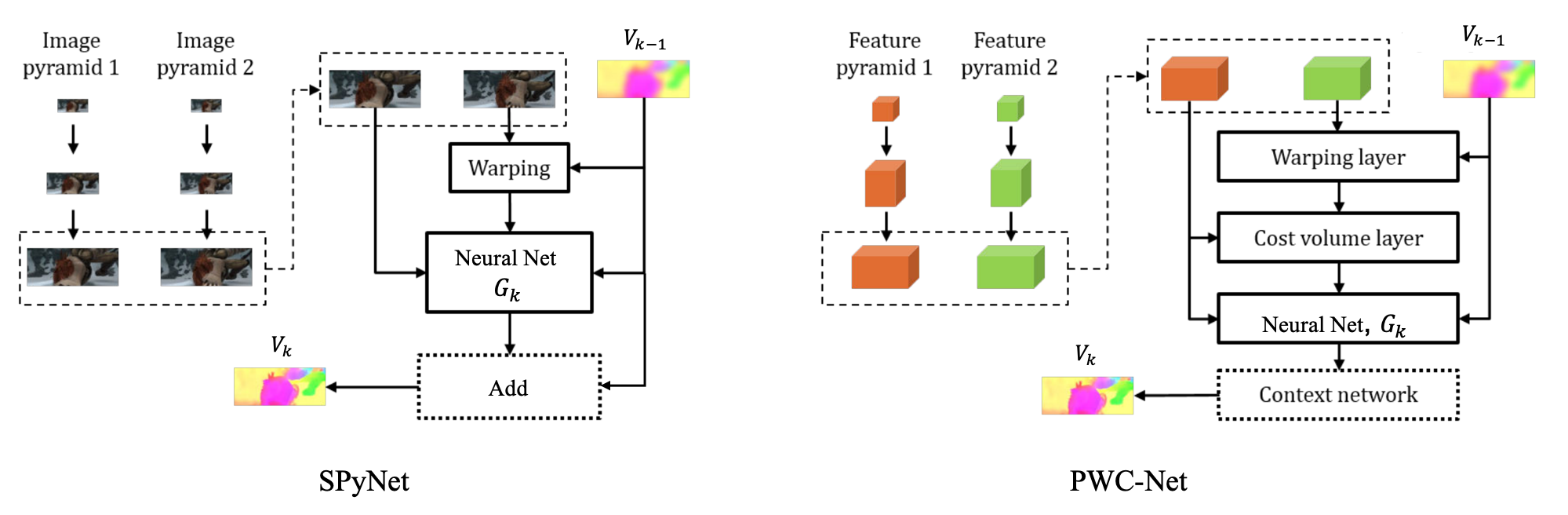}
}
\caption{\neww{Spatial Pyramid Network \cite{ranjan2016optical} (left)  and PWC-Net \cite{Sun2018PWC-Net} (right) for optical flow estimation. At each pyramid level, network $G_k$ predicts flow at that level which is used to condition the optical flow at the higher resolution level in the pyramid. Adapted from \cite{Sun2018PWC-Net}.}}
\label{fig:nets}
\end{figure*}

We train two different network architectures to estimate optical flow on both the \dataSingle and \dataMulti dataset. We choose \neww{compact models that are based on spatial pyramids, namely} SPyNet \cite{ranjan2016optical} and PWC-Net \cite{Sun2018PWC-Net}, shown in Figure \ref{fig:nets}. 
We \neww{denote} the models trained on the \dataSingle dataset by SPyNet+\dataSingle and PWC+\dataSingle. Similarly, we \neww{denote} models trained on the \dataMulti dataset by SPyNet+\dataMulti and PWC+\dataMulti. %

The spatial pyramid structure employs a convnet at each level of an image pyramid. A pyramid level works on a particular resolution of the image. The top level works on the full resolution and the image features are downsampled as we move to the bottom of the pyramid. Each level learns a convolutional layer $d$, to perform downsampling of image features. Similarly, a convolution layer $u$, is learned for decoding optical flow. At each level, the convnet $G_k$ predicts optical flow residuals $v_k$ at that level. These flow residuals get added at each level to produce the full flow, $V_K$ at the finest level of the pyramid.

In SPyNet, each convnet $G_k$ takes a pair of images as inputs along with flow $V_{k-1}$ obtained by \newww{resizing}
the output of the previous level \newww{with interpolation}. The second frame is however warped using $V_{k-1}$ and the triplet $\{I^1_k,  w(I^2_k, V_{k-1}),$ $  V_{k-1}\}$ is fed as input to the convnet $G_k$. 

\neww{In PWC-Net, a pair of image features, $\{I_k^1, I_k^2\}$ is input at a pyramid level, and the second feature map is warped using using the flow $V_{k-1}$ from the previous level of the pyramid. We then compute the cost-volume $c(I_k^1, w(I_k^2, V_{k-1}))$ over feature maps and pass it to network $G_k$ to compute optical flow $V_k$ at that pyramid level.}

We use the pretrained weights as initializations for training both SPyNet and PWC-Net. We train both  models end-to-end to minimize the average End Point Error (EPE).

{\textbf{Hyperparameters.}} 
\neww{We follow the same training procedure for SPyNet and PWC-Net. The only exception to this is the learning rate, which is determined empirically for each dataset and network from $\{10^{-6}, 10^{-5}, 10^{-4}\}$. 
For the \dataSingle we found $10^{-6}$ to yield best results for SpyNet.
Predictions of PWC on the \dataSingle dataset do not improve for any of these learning rates.
For training on \dataMulti a learning rate of $10^{-6}$ and $10^{-4}$ yield best results for SpyNet and PWC-Net, respectively.
We use Adam \cite{kingma2014adam} to optimize our loss with $\beta_1=0.9$ and $\beta_2=0.999$.} We use a batch size of $8$ and run $400,000$ training iterations. 
\neww{All networks are implemented in the Pytorch framework. Fine-tuning the networks from pretrained weights takes approximately 1 day on \dataSingle and 2 days on \dataMulti.
}

{\textbf{Data Augmentations.}} 
We also augment our data by applying several transformations and adding noise. Although our dataset is quite large, augmentation improves the quality of results on real scenes. In particular, we apply scaling in the range of $[0.3, 3]$, and rotations in $[-17^{\circ}, 17^{\circ}]$. 
The dataset is normalized to have zero mean and unit standard deviation using \cite{he2015deep}.

%% file: 5experiments.tex
\neww{In this section, we first compare the \dataSingle, \dataMulti and other common optical flow datasets. Next, we show that fine-tuning SPyNet on \dataSingle improves the model, while we observe that fine-tuning PWC-Net on \dataSingle does not improve the model further. We then fine-tune the same methods on \dataMulti and evaluate them. We show that both, SPyNet and PWC-Net improve when fine-tuned on \dataMulti. We show that the methods trained on the \dataMulti dataset outperform generic flow estimation methods for the pixels corresponding to humans.
We show on qualitative results that both, the models trained on \dataSingle and models trained on \dataMulti seem to generalize to real word scenes.}
\fin{Finally, we quantitatively evaluate optical flow methods on the \dataMulti dataset using motion compensated intensity metric.}

\textbf{Dataset Details.}
In comparison with other optical flow datasets, our dataset is larger by an order of magnitude (see Table \ref{tab:datasets});
the \dataSingle dataset contains $135,153$ training frames and $10,867$ test frames with optical flow ground truth, while
\new{the \dataMulti dataset has $86,259$ training, $13,236$ test and $11,817$ validation frames.
}
For the single-person dataset we keep the resolution small at $256 \times 256$ px to facilitate easy deployment for training neural networks.
This also speeds up the rendering process in Blender for generating large amounts of data.
We show the comparisons of processing time of different models on the \dataSingle dataset in Table \ref{tab:evaluation}(a).
\new{
For the \dataMulti dataset we increase the resolution to $640 \times 640$ px to be able to reason about optical flow even in small body parts like fingers, using SMPL+H.
}
Our data is extensive, containing a wide variety of human shapes, poses, actions and virtual backgrounds to support deep learning systems.
\linebreak

\begin{table}[t]
\centering
\begin{tabular}{lccc}
\toprule
Dataset          &\begin{tabular}[c]{@{}l@{}}{\small \# Train}\\ {\small Frames}\end{tabular} & \begin{tabular}[c]{@{}l@{}}{\small \# Test} \\ {\small Frames}\end{tabular} & {\small Resolution} \\ \midrule
{\small MPI Sintel~\cite{Butler:ECCV:2012}} & {\small $1,064  $}    &{\small $564 $ }    & {\small $1024 \times 436$ }\\
{\small KITTI 2012~\cite{Geiger2012CVPR}}      & {\small $194   $  }  & {\small $195 $ }    & {\small $1226 \times 370$} \\
{\small KITTI 2015~\cite{menze2015object}} & {\small $200 $ }   &{\small  $200 $ }    & {\small $1242 \times 375$} \\
{\small Virtual Kitti~\cite{Gaidon:Virtual:CVPR2016}}   & {\small $21,260 $ }   & $-   $    &{\small $1242 \times 375 $} \\
{\small Flying Chairs~\cite{dosovitskiy2015flownet}}   & {\small $22,232 $ }   &{\small $640 $  }   &{\small $512 \times 384 $} \\
{\small Flying Things~\cite{flyingthings} }& {\small $21,818 $}    & {\small $4,248$ }    & {\small $960 \times 540 $} \\
{\small Monkaa~\cite{flyingthings} }          & {\small $8,591  $ }   & $-   $     &{\small $960 \times 540 $} \\
{\small Driving~\cite{flyingthings}}         & {\small $4,392  $}    & $-   $     & {\small $960 \times 540 $} \\
{\small \dataSingle (ours)}       & {\small $135,153$ }   & {\small $ 10,867$}     & {\small $256 \times 256 $ }\\
{\small \dataMulti (ours)}       & {\small$86,259$ }   &{\small $ 13,236$}     & {\small$640 \times 640 $} \\ 
\bottomrule %
\end{tabular}
\vspace{0.02in}
\caption{Comparison of the {\dataSingleMultiFULL} datasets, namely the {\dataSingleFULL} (\dataSingle) and the {\dataMultiFULL} (\dataMulti) dataset, with previous optical flow datasets.}
\label{tab:datasets}
\end{table}

\begin{figure*}
  \centering
    \includegraphics[width=\textwidth]{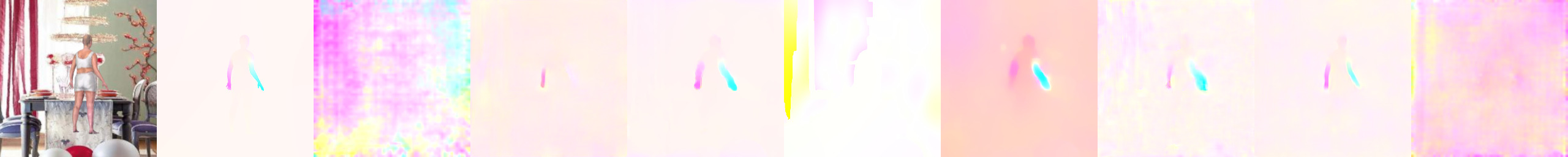}
    \includegraphics[width=\textwidth]{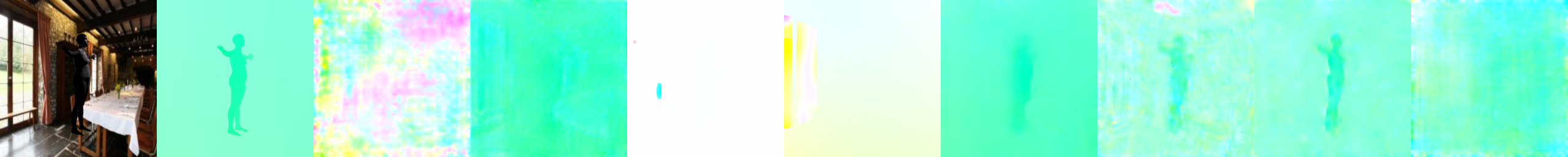}
    \includegraphics[width=\textwidth]{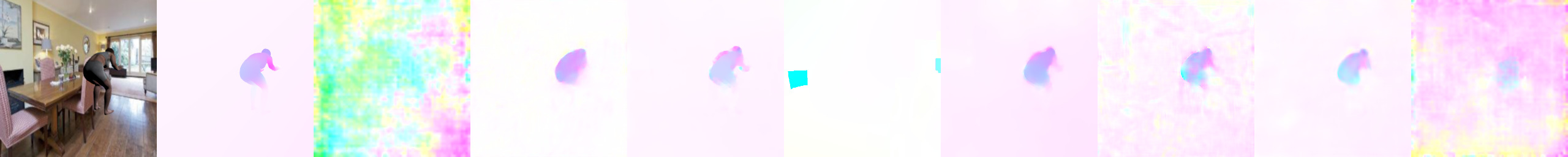}
    \includegraphics[width=\textwidth]{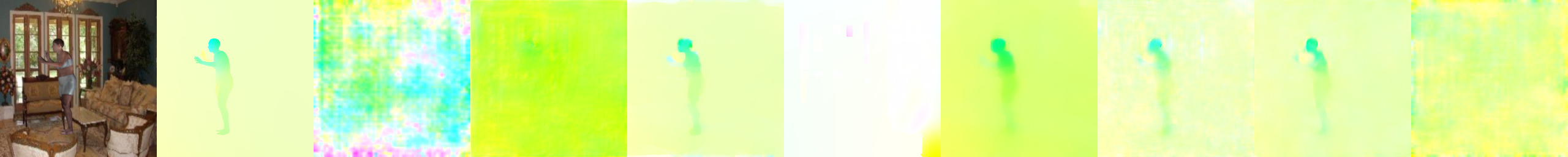}
    \includegraphics[width=\textwidth]{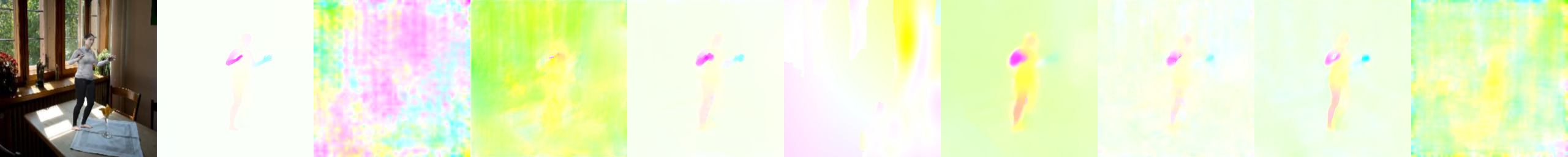}
    \includegraphics[width=\textwidth]{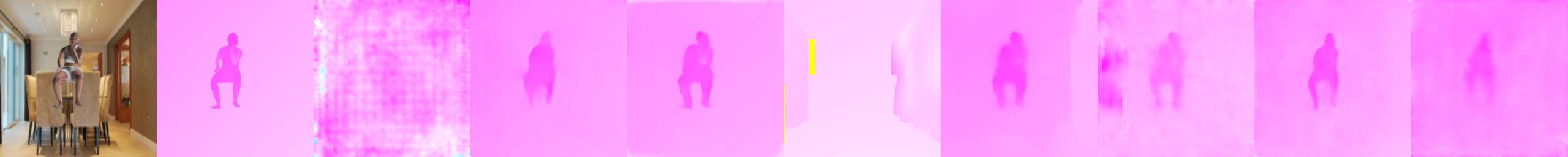}
    \includegraphics[width=\textwidth]{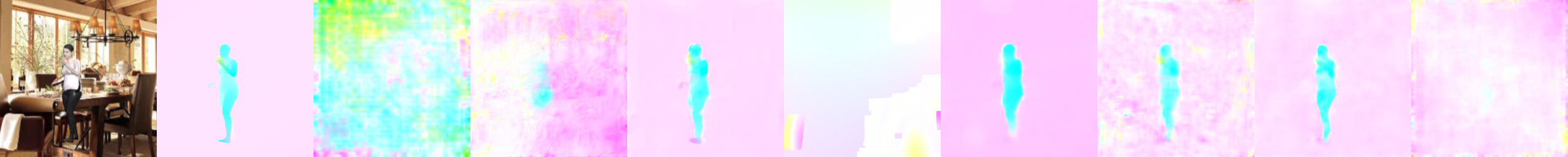}
    \includegraphics[width=\textwidth]{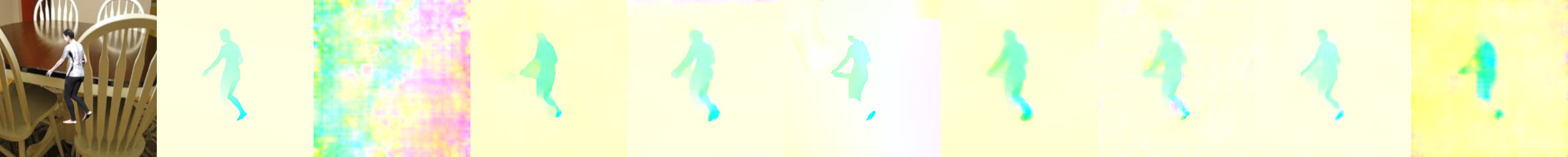}
\begin{flushleft}
\vspace{-0.2cm}
\quad \xspace Frame1   \quad Ground Truth  \qquad FlowNet \qquad FlowNet2 \qquad LDOF \qquad PCA-Layers \quad EpicFlow \qquad SPyNet \xspace SpyNet+\dataSingle \qquad PWC
\end{flushleft}

\vspace{-0.1in}
\caption{Visual comparison of optical flow estimates using different methods on the \dataSingleFULL (\dataSingle) test set. From left to right, we show Frame 1, Ground Truth flow, results of \neww{FlowNet} \cite{dosovitskiy2015flownet}, FlowNet2 \cite{ilg2016flownet}, LDOF \cite{brox2009large}, PCA-Layers \cite{wulff2015efficient}, SPyNet \cite{ranjan2016optical}, EpicFlow \cite{epicflow} , LDOF \cite{brox2009large}, SPyNet \cite{ranjan2016optical}, SPyNet+\dataSingle (ours) and PWC-Net \cite{Sun2018PWC-Net}
}
\label{fig:results}
\end{figure*}

\textbf{Comparison on \dataSingle.}
We compare the average End Point Errors (EPEs) of optical flow methods \neww{on the \dataSingle dataset} in \neww{Table \ref{tab:evaluation}}, along with the time for evaluation.
We show visual comparisons in Figure \ref{fig:results}.
Human motion is complex and general optical flow methods fail to capture it. 
\fin{We observe that SPyNet+\dataSingle outperforms methods that are not trained on \dataSingle, and SPyNet \cite{ranjan2016optical} in particular.} 
\fin{We expect more involved methods like \neww{FlowNet2}~\cite{ilg2016flownet} to have bigger performance gain than SPyNet when trained on \dataSingle.} 

We observe that \neww{FlowNet} \cite{dosovitskiy2015flownet} shows poor generalization on our dataset. Since the results of \neww{FlowNet} \cite{dosovitskiy2015flownet} in Table \ref{tab:evaluation} and \ref{tab:partsEPE} are very close to the zero flow (no motion) baseline, we cross-verify by evaluating \neww{FlowNet} on a mixture of Flying Chairs \cite{dosovitskiy2015flownet} and \dataSingleMultiFULL and observe that the flow outputs on \dataSingle is quite random (see Figure \ref{fig:results}). The main reason is that \dataSingle contains a significant amount of small motions and it is known that \neww{FlowNet} does not perform very well on small motions. SPyNet+SHOF \cite{ranjan2016optical} however performs quite well and is able to generalize to body motions. The results however look noisy in many cases.

Our dataset employs a layered structure where a human is placed against a background. As such layered methods like PCA-layers \cite{wulff2015efficient} perform very well on a few images (row 8 in Figure \ref{fig:results}) where they are able to segment a person from the background. However, in most cases, they do not obtain good segmentation into layers.

Previous state-of-the-art methods like LDOF \cite{brox2009large} and  Epic-Flow \cite{epicflow} perform much better than others. They get a good overall shape, and smooth backgrounds. However, their estimation is quite blurred. They tend to miss the sharp edges that are typical of human hands and legs. They are also significantly slower.

In contrast, \neww{by fine-tuning on our dataset, the performance of SPyNet+\dataSingle improves by 40\%} {over SPyNet} on the \dataSingle dataset.
\fin{We also find that fine-tuning PWC-Net on the \dataSingle does not improve the model. This could be because SHOF dataset has predominantly small motion which is handled better by SPyNet~\cite{ranjan2016optical} architecture.
Empirically, we have seen that PWC-Net has state-of-the-art performance on standard benchmarks.
This motivates the generation of the \dataMulti dataset, which includes larger motions and more complex scenes with occlusions.}

\fin{A qualitative comparison to popular optical flow methods can be seen in Figure~\ref{fig:results}. Flow estimations of SPyNet+\dataSingle can be observed to be sharper than those of methods that are not trained on human motion. This can especially be seen for edges. 
}

\begin{table}[t]
\centering
\resizebox{0.48\textwidth}{!}{
\begin{tabular}{lcccc}
\toprule
Method     & {AEPE} & {Time(s)} & \neww{Learned} & \neww{Fine-tuned on } \\
 & & & &\neww{\dataSingle} \\ \midrule
{Zero} & {0.6611} &  - & -  \\ \midrule
{FlowNet~\cite{dosovitskiy2015flownet} } & {0.5846}          &  {0.080}          &   \cmark   &  \xmark   \\
{PCA Layers~\cite{wulff2015efficient}}     &{0.3652}           & {10.357}          &   \xmark   &  \xmark                \\
\neww{{PWC-Net \cite{Sun2018PWC-Net}}}  & \neww{0.2158}     & \neww{0.024}     &   \cmark   &  \xmark    \\ %
\neww{{PWC+\dataSingle}}  & \neww{0.2158}      &\neww{0.024}     &   \cmark   &  \cmark   \\

{SPyNet~\cite{ranjan2016optical} }       &{0.2066}           & \textbf{\neww{0.022}}           &   \cmark   &  \xmark    \\
{Epic Flow~\cite{epicflow}}              &{0.1940}           & {1.863}           &   \xmark   &  \xmark                \\
{LDOF~\cite{brox2009large}}              & {0.1881}          & {8.620}           &   \xmark   &  \xmark                \\
\neww{FlowNet2}~\cite{ilg2016flownet} & \neww{0.1895} &  \neww{0.127} & \cmark & \xmark\\
{Flow Fields~\cite{flowfields} }         & {0.1709}          & {4.204}           &   \xmark   &  \xmark                \\
{SPyNet+\dataSingle}                           &{\textbf{0.1164}}  & \textbf{\neww{0.022}}  &   \cmark   &  \cmark	    \\

\bottomrule
\end{tabular}
}
\vspace{0.02in}
\caption{
EPE comparisons and evaluation times of different optical flow methods on the \dataSingle dataset. Zero refers to the EPE when zero flow (no motion) is always used for evaluation. \new{Evaluation times are based on the \dataSingle dataset with $256 \times 256$ image resolution}. \neww{We time all GPU based methods using a Tesla V100-16GB GPU}.}

\label{tab:evaluation}
\end{table}

\begin{figure*}
  \centering
    \includegraphics[width=\textwidth]{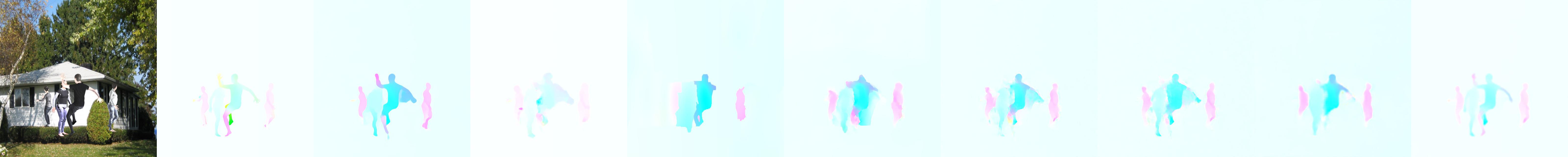}
    \includegraphics[width=\textwidth]{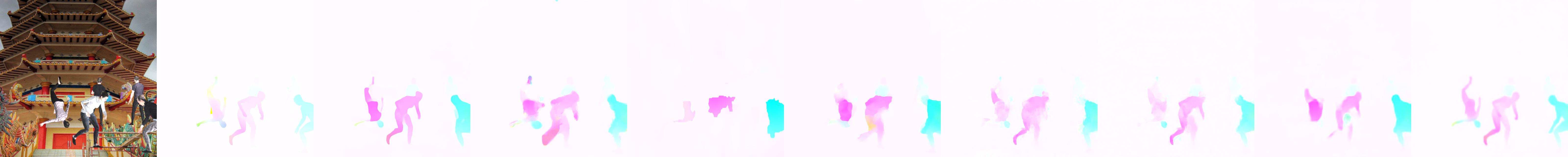}
    \includegraphics[width=\textwidth]{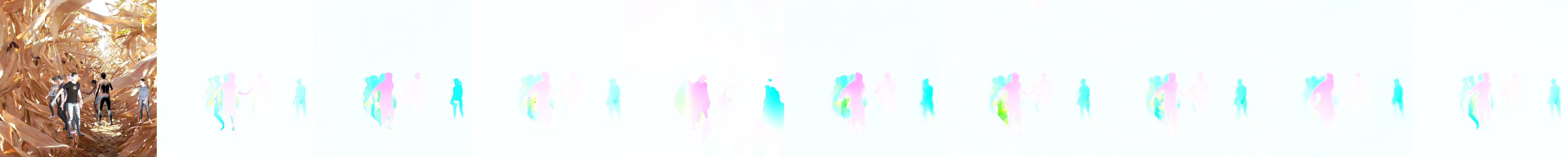}
    \includegraphics[width=\textwidth]{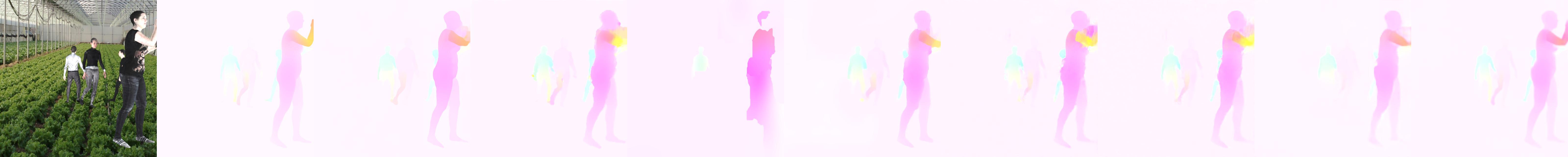}
   \includegraphics[width=\textwidth]{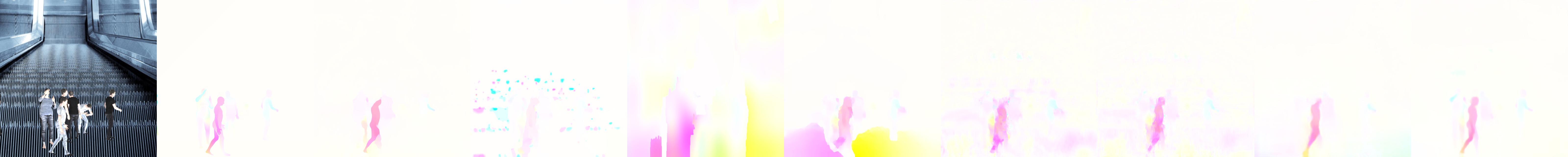}
    \includegraphics[width=\textwidth]{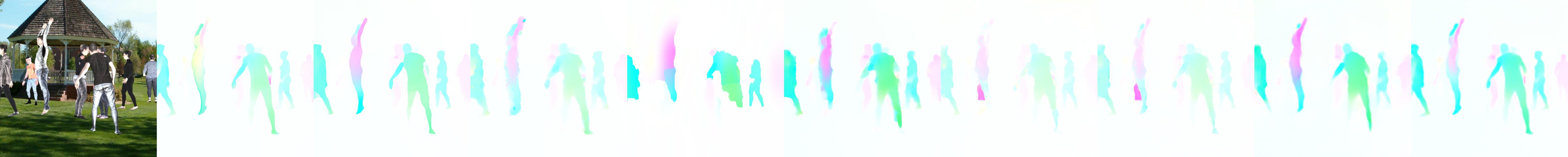}
    \includegraphics[width=\textwidth]{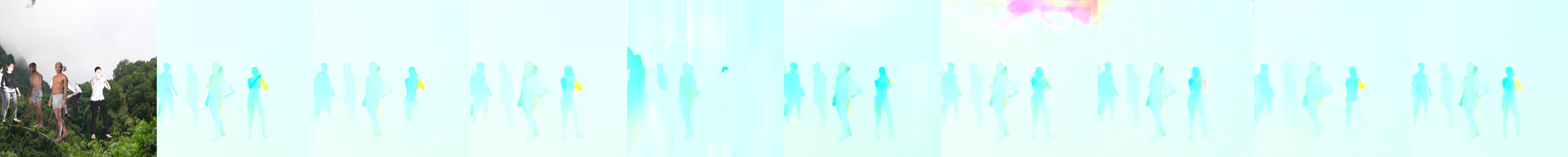}

\begin{tabular}{C{0.066\textwidth}C{0.1\textwidth}C{0.07\textwidth}C{0.06\textwidth}C{0.09\textwidth}C{0.07\textwidth}C{0.06\textwidth}C{0.113\textwidth}C{0.04\textwidth}C{0.098\textwidth}}
Frame1 & Ground Truth & FlowNet2 & LDOF & PCA-Layers & EpicFlow &SPyNet & SPyNet+\dataMulti &PWC &PWC+\dataMulti\\
\end{tabular}
\caption{Visual comparison of optical flow estimates using different methods on the \dataMultiFULL (\dataMulti) test set. From left to right, we show Frame 1, Ground Truth flow, results of FlowNet2 \cite{ilg2016flownet}, LDOF \cite{brox2009large}, PCA-Layers \cite{wulff2015efficient}, EpicFlow \cite{epicflow}, SPyNet \cite{ranjan2016optical}, SPyNet+\dataMulti (ours), PWC-Net \cite{Sun2018PWC-Net} and PWC+\dataMulti (ours).
}
\label{fig:results_multi}
\end{figure*}

\textbf{Comparison on \dataMulti.}
\neww{ Training (fine-tuning) on the \dataMulti dataset improves SPyNet and PWC-Net on average, as can be seen in Table~\ref{tab:epeMulti}.
In particular PWC+\dataMulti outperforms \mbox{SPyNet+\dataMulti} and also improves over generic state-of-the-art optical flow methods.
Large parts of the image are background, whose movements are relatively easy to estimate. However, we are particularly interested in human motions. Therefore, we mask out all errors of background pixels and compute the average EPE only on body pixels (see Table~\ref{tab:epeMulti}).
For these pixels, light-weight networks like SpyNet and PWC-Net improve over almost all generic optical flow estimation methods using our dataset (SpyNet+\dataMulti and PWC+\dataMulti), including the much larger network FlowNet2. PWC+\dataMulti is the best performing method.}

\neww{A more fine grained analysis of EPE across body parts is shown in Table \ref{tab:partsEPE}. We obtain EPE of these body parts using the segmentation shown in Figure \ref{fig:segmentation}. It can be seen that improvements of PWC+\dataMulti over FlowNet2 are larger for body parts that are at the end of the kinematic tree (i.e. feet, calves, arms and in particular fingers). Differences are less strong for body parts close to the torso. One interpretation of these findings is that movements of the torso are easier to predict, while movements of body parts at the end of the kinematic tree are more complex and thus harder to estimate.
In contrast, SPyNet+\dataMulti outperforms FlowNet2 on body parts close to the torso and does not learn to capture the more complex motions of limbs better than FlowNet2.}
\fin{We expect FlowNet2+\dataMulti to perform even better, but we do not include this here due to its long and tedious training process.}

\begin{table}[t]
\centering
\begin{tabular}{lccc}
\toprule
Method &Average  & Average EPE on  & Fine-tuned on \\
 & EPE &  \emph{body pixels}   & \dataMulti\\
\midrule
FlowNet &0.808 & 2.574 &  \xmark\\
PCA Layers &0.556 & 2.691 & \xmark\\
Epic Flow & 0.488 & 1.982 & \xmark\\
SPyNet & 0.429 & 1.977&  \xmark\\
SPyNet+MHOF &0.391 & 1.803 & \cmark\\
PWC-Net & 0.369 & 2.056&  \xmark\\
LDOF & 0.360 & 1.719 &\xmark\\
FlowNet2 &0.310 & 1.863 &\xmark\\
PWC+MHOF &\textbf{0.301} & \textbf{1.621}  &\cmark \\
\bottomrule
\end{tabular}
\vspace{0.02in}
\caption{
\neww{Comparison using End Point Error (EPE) on the \dataMultiFULL (\dataMulti) dataset. We show the average EPE and body-only EPE. %
\newww{For the latter,} the EPE is computed only over segments of the image depicting a human body. Best results are shown in boldface.
A comparison of body-part specific EPE can be found in Table~\ref{tab:partsEPE} }.
}

\label{tab:epeMulti}
\end{table}

\begin{table}[t]
    \centering
    \begin{tabular}{lcc}
        \toprule
        Method  & Average  MCI & Average MCI \\
                & \dataMulti   & Real \\
        \midrule
        FlowNet & 287.328 & 401.779 \\
        PCA Layers & 201.594 &  423.332\\
        Epic Flow & 129.252 & 234.037\\
        SPyNet &  142.108 &  302.753\\
        SPyNet+MHOF & 143.029 & 297.142\\
        PWC-Net & 157.088 & 344.202 \\
        LDOF & \textbf{71.449} & \textbf{158.281} \\
        FlowNet2 & 145.732  & 303.799 \\
        PWC+MHOF & 152.314 & 351.567 \\
        \bottomrule
    \end{tabular}
    \vspace{0.02in}
    \caption{
    \newww{Comparison using Motion Compensated Intensity (MCI) on the \dataMultiFULL (\dataMulti) dataset and a real video sequence. Example images for the real video sequence can be seen in Figure \ref{fig:results_real_multi}.}
    }
    \label{tab:MCI}
\end{table}

\setlength{\tabcolsep}{5pt}
\begin{table*}[t]
\centering
\resizebox{!}{0.55\textwidth}{
    \begin{tabular}{lccccccccc}
    \toprule
    Parts &Epic Flow & LDOF & FlowNet2 & FlowNet & PCA Layers & PWC-Net & PWC+MHOF & SPyNet & SPyNet+MHOF\\
    \midrule
    Average (whole image) & 0.488 & 0.360 & 0.310 & 0.808 & 0.556 & 0.369   & \textbf{0.301} & 0.429 & 0.391 \\
    Average (body pixels) & 1.982 & 1.719 & 1.863 & 2.574 & 2.691 & 2.056   & \textbf{1.621} & 1.977 & 1.803 \\
    \midrule
    global & 1.269 & 1.257 & 1.337 & 2.005 & 1.920 & 1.389                  & \textbf{1.163} & 1.356 & 1.236 \\
    head & 1.806 & \textbf{1.328} & 1.626 & 2.681 & 2.808 & 1.881           & 1.445 & 1.708 & 1.519 \\
    leftCalf & 2.116 & 1.802 & 1.787 & 2.420 & 2.711 & 2.109                & \textbf{1.476} & 1.991 & 1.796 \\
    leftFoot & 3.089 & 2.346 & 2.476 & 2.987 & 3.393 & 3.002                & \textbf{2.142} & 2.701 & 2.566 \\
    leftForeArm & 3.972 & 3.231 & 3.536 & 4.380 & 4.778 & 3.926             & \textbf{3.136} & 3.945 & 3.605 \\
    leftHand & 5.777 & 4.422 & 4.823 & 5.928 & 6.531 & 5.634                & \textbf{4.385} & 5.547 & 5.040 \\
    leftShoulder & 1.513 & \textbf{1.429} & 1.646 & 2.331 & 2.336 & 1.732   & 1.471 & 1.560 & 1.462 \\
    leftThigh & 1.424 & 1.338 & 1.466 & 2.102 & 2.150 & 1.565               & \textbf{1.230} & 1.517 & 1.362 \\
    leftToes & 3.147 & 2.573 & 2.755 & 3.065 & 3.307 & 3.100                & \textbf{2.524} & 2.830 & 2.784 \\
    leftUpperArm & 2.215 & \textbf{1.947} & 2.288 & 3.005 & 3.139 & 2.376   & 1.955 & 2.307 & 2.076 \\
    lIndex0 & 6.199 & 4.900 & 5.334 & 6.254 & 6.785 & 6.124                 & \textbf{4.861} & 5.925 & 5.472 \\
    lIndex1 & 6.367 & \textbf{5.159} & 5.672 & 6.340 & 6.829 & 6.303        & 5.212 & 6.087 & 5.727 \\
    lIndex2 & 6.315 & \textbf{5.253} & 5.878 & 6.203 & 6.670 & 6.270        & 5.433 & 6.028 & 5.784 \\
    lMiddle0 & 6.338 & 4.983 & 5.331 & 6.364 & 6.910 & 6.211                & \textbf{4.837} & 6.012 & 5.544 \\
    lMiddle1 & 6.498 & 5.239 & 5.632 & 6.435 & 6.927 & 6.383                & \textbf{5.176} & 6.143 & 5.767 \\
    lMiddle2 & 6.266 & \textbf{5.212} & 5.756 & 6.130 & 6.592 & 6.182       & 5.303 & 5.934 & 5.679 \\
    lPinky0 & 6.048 & \textbf{4.792} & 5.302 & 6.035 & 6.603 & 5.940        & 4.873 & 5.738 & 5.307 \\
    lPinky1 & 6.106 & \textbf{4.922} & 5.489 & 6.038 & 6.574 & 6.014        & 5.064 & 5.765 & 5.418 \\
    lPinky2 & 5.780 & \textbf{4.856} & 5.419 & 5.655 & 6.170 & 5.702        & 4.956 & 5.474 & 5.231 \\
    lRing0 & 6.388 & 4.973 & 5.281 & 6.413 & 7.010 & 6.218                  & \textbf{4.834} & 6.064 & 5.552 \\
    lRing1 & 6.313 & 5.083 & 5.391 & 6.256 & 6.801 & 6.168                  & \textbf{4.949} & 5.966 & 5.558 \\
    lRing2 & 6.047 & \textbf{5.035} & 5.515 & 5.924 & 6.409 & 5.942         & 5.067 & 5.710 & 5.441 \\
    lThumb0 & 5.415 & \textbf{4.318} & 4.673 & 5.473 & 6.072 & 5.316        & 4.329 & 5.212 & 4.809 \\
    lThumb1 & 5.636 & \textbf{4.527} & 5.065 & 5.698 & 6.232 & 5.612        & 4.685 & 5.449 & 5.065 \\
    lThumb2 & 5.825 & \textbf{4.749} & 5.388 & 5.820 & 6.323 & 5.802        & 5.005 & 5.629 & 5.314 \\
    neck & 1.336 & \textbf{1.195} & 1.371 & 2.151 & 2.245 & 1.440           & 1.227 & 1.399 & 1.250 \\
    rightCalf & 2.243 & 1.892 & 1.864 & 2.530 & 2.851 & 2.223               & \textbf{1.548} & 2.081 & 1.907 \\
    rightFoot & 3.270 & 2.454 & 2.610 & 3.149 & 3.599 & 3.171               & \textbf{2.276} & 2.894 & 2.732 \\
    rightForeArm & 3.990 & 3.242 & 3.554 & 4.381 & 4.759 & 3.928            & \textbf{3.190} & 4.029 & 3.641 \\
    rightHand & 5.735 & 4.348 & 4.787 & 5.837 & 6.447 & 5.550               & \textbf{4.339} & 5.582 & 4.978 \\
    rightShoulder & 1.547 & \textbf{1.431} & 1.670 & 2.390 & 2.340 & 1.735  & 1.477 & 1.573 & 1.462 \\
    rightThigh & 1.477 & 1.374 & 1.512 & 2.158 & 2.226 & 1.624              & \textbf{1.263} & 1.556 & 1.407 \\
    rightToes & 3.395 & 2.707 & 2.918 & 3.293 & 3.566 & 3.346               & \textbf{2.699} & 3.064 & 2.999 \\
    rightUpperArm & 2.267 & \textbf{1.974} & 2.294 & 3.033 & 3.148 & 2.400  & 2.007 & 2.002 & 2.113 \\
    rIndex0 & 6.264 & \textbf{4.875} & 5.324 & 6.255 & 6.800 & 6.150        & 4.886 & 6.003 & 5.486 \\
    rIndex1 & 6.541 & \textbf{5.210} & 5.755 & 6.449 & 6.951 & 6.457        & 5.329 & 6.237 & 5.835 \\
    rIndex2 & 6.465 & \textbf{5.320} & 5.968 & 6.294 & 6.776 & 6.404        & 5.533 & 6.149 & 5.879 \\
    rMiddle0 & 6.509 & 5.056 & 5.454 & 6.470 & 7.014 & 6.354                & \textbf{4.967} & 6.211 & 5.662 \\
    rMiddle1 & 6.680 & 5.341 & 5.777 & 6.562 & 7.058 & 6.537                & \textbf{5.325} & 6.325 & 5.895 \\
    rMiddle2 & 6.394 & \textbf{5.261} & 5.838 & 6.209 & 6.713 & 6.274       & 5.366 & 6.038 & 5.739 \\
    rPinky0 & 5.983 & \textbf{4.750} & 5.372 & 5.952 & 6.504 & 5.855        & 4.845 & 5.741 & 5.262 \\
    rPinky1 & 6.076 & \textbf{4.905} & 5.566 & 5.979 & 6.533 & 5.943        & 5.025 & 5.809 & 5.402 \\
    rPinky2 & 5.789 & \textbf{4.813} & 5.403 & 5.645 & 6.220 & 5.662        & 4.903 & 5.532 & 5.232 \\
    rRing0 & 6.397 & 4.948 & 5.350 & 6.383 & 6.938 & 6.215                  & \textbf{4.856} & 6.126 & 5.565 \\
    rRing1 & 6.395 & 5.108 & 5.465 & 6.290 & 6.841 & 6.212                  & \textbf{5.019} & 6.066 & 5.615 \\
    rRing2 & 6.222 & \textbf{5.129} & 5.644 & 6.052 & 6.610 & 6.063         & 5.160 & 5.889 & 5.571 \\
    rThumb0 & 5.417 & \textbf{4.304} & 4.748 & 5.470 & 6.057 & 5.301        & 4.360 & 5.247 & 4.819 \\
    rThumb1 & 5.605 & \textbf{4.465} & 4.945 & 5.643 & 6.210 & 5.514        & 4.607 & 5.434 & 5.032 \\
    rThumb2 & 5.835 & \textbf{4.748} & 5.262 & 5.789 & 6.328 & 5.749        & 4.938 & 5.639 & 5.306 \\
    spine & 1.233 & 1.271 & 1.325 & 1.941 & 1.856 & 1.360                   & \textbf{1.168} & 1.322 & 1.221 \\
    spine1 & 1.330 & 1.369 & 1.421 & 2.028 & 1.957 & 1.460                  & \textbf{1.268} & 1.417 & 1.322 \\
    spine2 & 1.329 & 1.308 & 1.439 & 2.089 & 2.049 & 1.480                  & \textbf{1.276} & 1.387 & 1.309 \\
    \bottomrule
    \end{tabular}
}
\vspace{0.02in}
\caption{Comparison using End Point Error (EPE) on the \dataMultiFULL (\dataMulti) dataset. We show the average EPE and body part specific EPE, where part labels follow Figure \ref{fig:segmentation}. %
The first two rows are repeated from Tab~\ref{tab:epeMulti}.}
\label{tab:partsEPE}
\end{table*}

\begin{figure}
\centering     %
\includegraphics[width=0.45\textwidth]{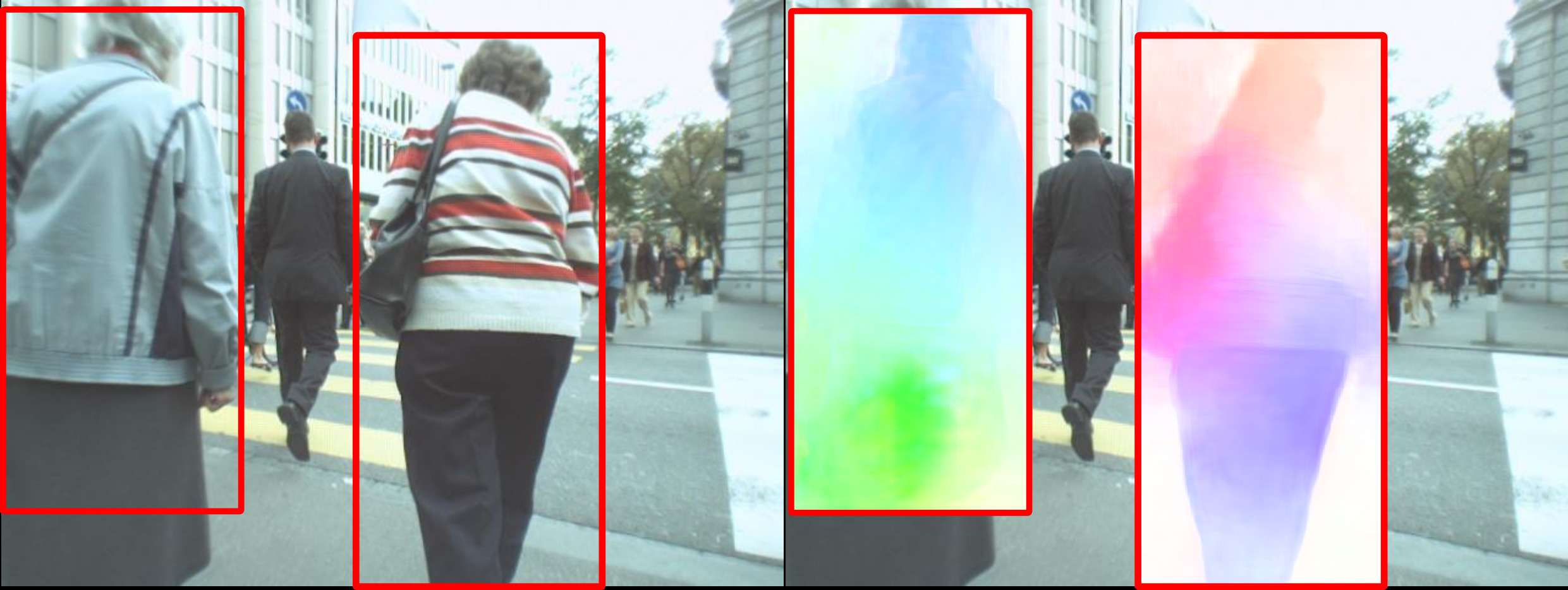}
\caption{We use \neww{the DPM \cite{Felzenszwalb2010PAMI}} person detector to crop out people from real-world scenes (left) and use SPyNet+\dataSingle to compute optical flow on the cropped section (right).}
\label{fig:persondetect}
\end{figure}

\neww{Visual comparisons are shown in Figure \ref{fig:results_multi}. In particular, PWC+\dataMulti predicts flow fields with sharper edges than generic methods or SPyNet+ \dataMulti.
Furthermore, the qualitative results suggest that PWC+\dataMulti is better at distinguishing the motion of people, as people can be better separated on the flow visualizations of PWC+\dataMulti (Figure~\ref{fig:results_multi}, row 3).
Last, it can be seen that fine details, like the motion of distant humans or small body parts, are better estimated by PWC+\dataMulti.}

\neww{The above observations are strong indications that our \dataSingleMultiFULL datasets (\dataSingle and \dataMulti) can be beneficial for the performance on human motion for other optical flow networks as well.}

\begin{figure*}
  \centering
    \includegraphics[width=\textwidth]{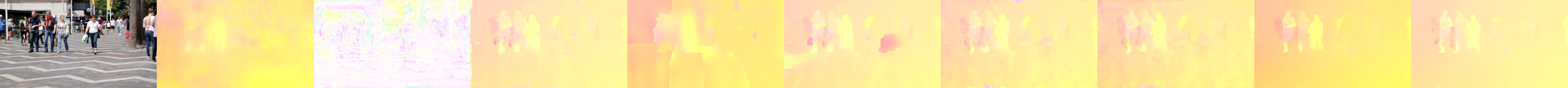}
    \includegraphics[width=\textwidth]{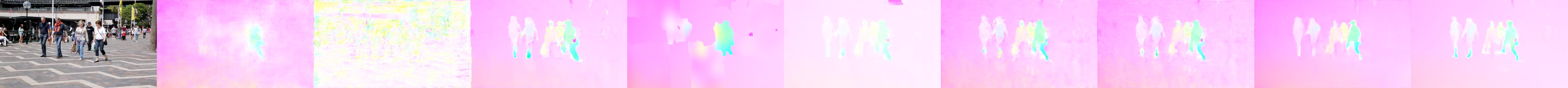}
    \includegraphics[width=\textwidth]{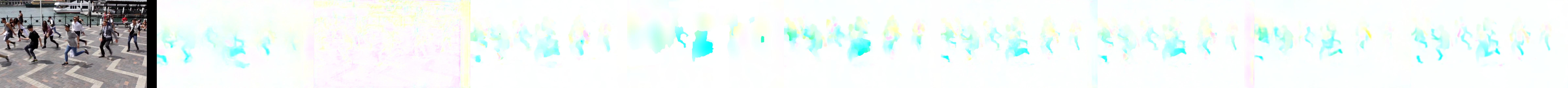}
   \includegraphics[width=\textwidth]{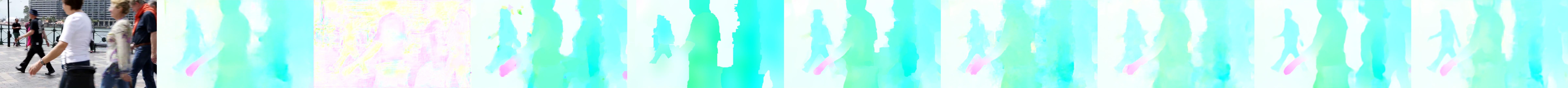}
    \includegraphics[width=\textwidth]{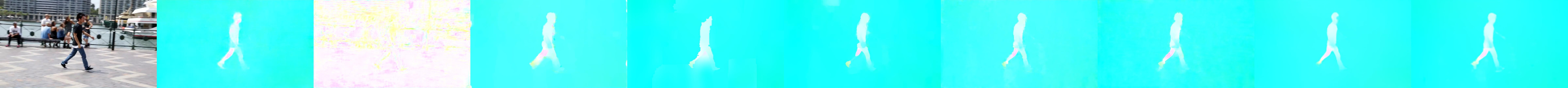}
    \includegraphics[width=\textwidth]{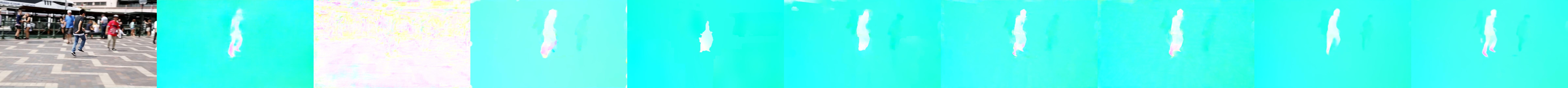}
    \includegraphics[width=\textwidth]{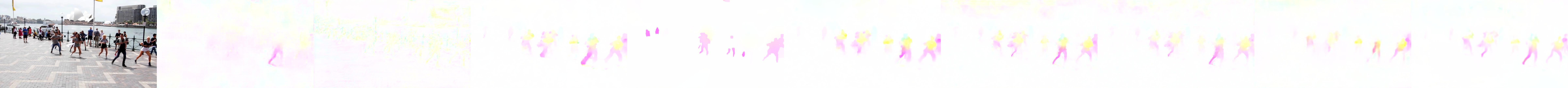}
    \includegraphics[width=\textwidth]{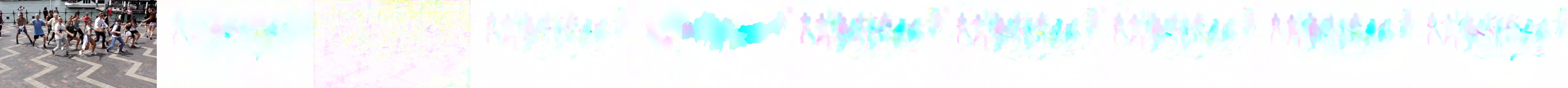}

\begin{tabular}{C{0.066\textwidth}C{0.09\textwidth}C{0.08\textwidth}C{0.07\textwidth}C{0.09\textwidth}C{0.07\textwidth}C{0.053\textwidth}C{0.115\textwidth}C{0.038\textwidth}C{0.098\textwidth}}
Frame1 & FlowNet2 & FlowNet & LDOF & PCA-Layers & EpicFlow &SPyNet & SPyNet+\dataMulti &PWC &PWC+\dataMulti\\
\end{tabular}
\vspace{-0.1in}
\caption{\dataMultiFULL visuals on real images. From left to right, we show Frame 1, results of FlowNet2 \cite{ilg2016flownet}, FlowNet \cite{dosovitskiy2015flownet}, LDOF \cite{brox2009large}, PCA-Layers \cite{wulff2015efficient}, EpicFlow \cite{epicflow}, SPyNet \cite{ranjan2016optical}, SPyNet+\dataMulti(ours), PWC-Net \cite{Sun2018PWC-Net} and PWC+\dataMulti (ours).
}
\label{fig:results_real_multi}
\end{figure*}

\textbf{Real Scenes.}
\begin{figure*}
  \centering
    \includegraphics[width=\textwidth, trim={0 1cm 0 0},clip]{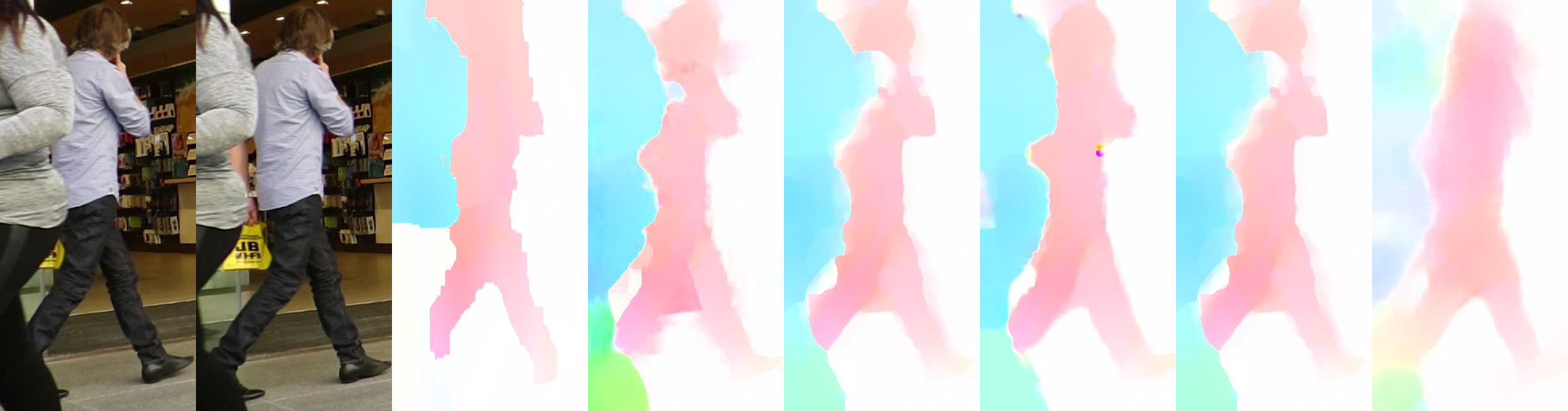}
    \includegraphics[width=\textwidth, trim={0 0 0 1.5cm},clip]{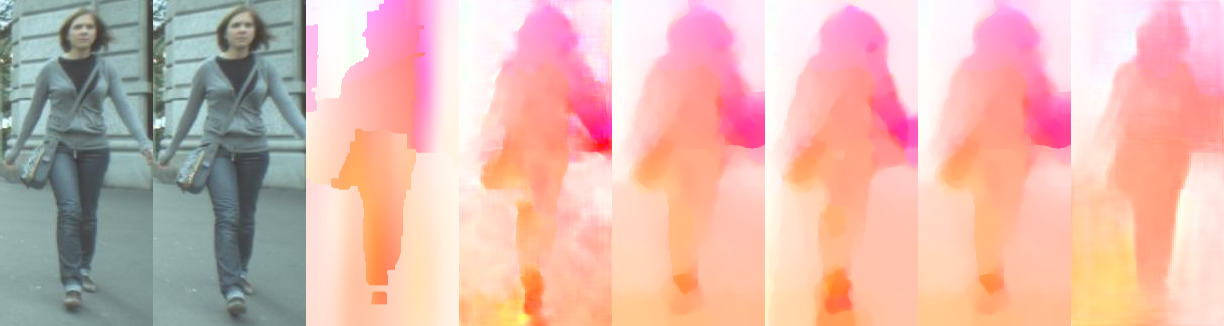}

    \includegraphics[width=\textwidth]{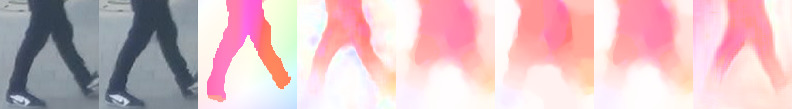}
    \includegraphics[width=\textwidth]{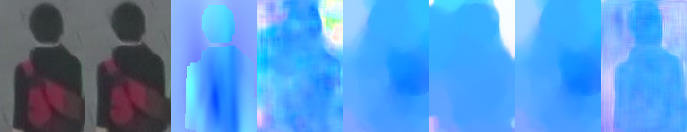}
\vspace{-0.2in}
\begin{flushleft}
\hspace{0.3in} {Frame 1} \hspace{0.4in} {Frame 2}\hspace{0.3in} {PCA-Layers} \hspace{0.3in}{SPyNet}\hspace{0.4in} {EpicFlow} \hspace{0.45in} {LDOF}\hspace{0.38in} {FlowFields}\hspace{0.29in}{SPyNet+\dataSingle}
\end{flushleft}
\caption{
\neww{\dataSingleFULL visuals} on real images using different methods.
From left to right, we show Frame 1, Frame 2, results of PCA-Layers \cite{wulff2015efficient}, and SPyNet \cite{ranjan2016optical}, EpicFlow \cite{epicflow}, LDOF \cite{brox2009large}, FlowFields \cite{flowfields} and SPyNet+\dataSingle (ours). }%

\label{fig:realpeople}
\end{figure*}
We show a visual comparison of results on real-world scenes of people in motion. \neww{For visual comparisons of models trained on the \dataSingle dataset} we collect these scenes by cropping people from real world videos as shown in Figure \ref{fig:persondetect}. We use DPM \cite{Felzenszwalb2010PAMI} for detecting people and compute bounding box regions in two frames using the ground truth of the MOT16 dataset \cite{MilanL0RS16}. The results for the \dataSingle dataset are shown in Figure~\ref{fig:realpeople}. \neww{A comparison of methods on real images with multiple people can be seen in Figure~\ref{fig:results_real_multi}.}

The performance of PCA-Layers \cite{wulff2015efficient} is highly dependent on its ability to segment. Hence, we see only a few cases where it looks visually correct. SPyNet \cite{ranjan2016optical} gets the overall shape but the results look noisy in certain image parts. While LDOF \cite{brox2009large}, EpicFlow \cite{epicflow} and FlowFields \cite{flowfields} generally perform  well, they often find it difficult to resolve the legs, hands and head of the person.
The results from models trained on our \dataSingleMultiFULL dataset look appealing especially while resolving the overall human shape, and various parts like legs, hands and the human head.
\neww{Models trained on the \dataSingleMultiFULL dataset perform well under occlusion (Figure~\ref{fig:realpeople}, Figure \ref{fig:results_real_multi}). Many examples including severe occlusion can be seen in Figure~\ref{fig:results_real_multi}.
Besides that, Figure~\ref{fig:results_real_multi} shows that the models trained on \dataMulti are able to distinguish motions of multiple people and predict sharp edges of humans.}

\fin{A quantitative evaluation on real data with humans is not possible, as no such dataset with ground truth optical flow annotation exists. 
To determine generalization of the models to real data, despite the lack of ground truth annotation, we can use the Motion Compensated Intensity (MCI) as an error metric. Given the image sequence $I^1, I^2$ and predicted flow $V$, the MCI error is given by}

\begin{align}
\qquad \qquad \text{MCI}(I^1, I^2, V) &= ||I^1 - w(I^2, V)||^2,
\end{align}
\fin{where $w$ warps the image $I^2$ according to flow $V$. This metric certainly has limitations. The motion compensated intensity assumes Lambertian conditions i.e. intensity of a point remains constant over time. MCI error does not account for occlusions. Furthermore, MCI does not account for smooth flow fields over texture-less surfaces. 
Despite these shortcoming of MCI, we report these numbers to show that our models generalize to real data. However, it should be noted that EPE is a more precise metric to evaluate optical flow estimation.
} 

\fin{
To test whether MCI correlates with EPEs in Table \ref{tab:epeMulti}, we compute MCI on the \dataMulti dataset. 
The results can be seen in Table \ref{tab:MCI}. We observe that, methods like FlowNet and PCA-Layers which have poor performance on the EPE metric have higher MCI error. 
For methods with lower EPE, the MCI errors do not exactly correspond to the respective EPEs. 
This is due to the limitations of the MCI metric, as described above.
Finally, we compute MCI on a real video sequence from Youtube\footnote{\url{https://www.youtube.com/watch?v=2DiQUX11YaY}}. The MCI errors are shown in Table~\ref{tab:MCI}.
}

%% file: 6conclusion_javier.tex
In summary, we created an extensive \dataSingleMultiFULL dataset containing images of realistic human shapes in motion together with ground truth optical flow.
The dataset is comprised of two parts, the \dataSingleFULL (\dataSingle) and the \dataMultiFULL (\dataMulti) dataset.
We then train two compact network architectures based on spatial pyramids, namely SpyNet and PWC-Net.
The realism and extent of our dataset, together with an end-to-end training scheme, allows these networks to outperform previous state-of-the-art optical flow methods on our new human-specific dataset.
This indicates that our dataset can be beneficial for other optical flow network architectures as well.
Furthermore, our qualitative results suggest that the networks trained on the \dataSingleMultiFULL generalize well to real world scenes with humans.
\fin{This is evidenced by results on a real sequence using the MCI metric.}
The trained models are compact and run in real time making them highly suitable for phones and embedded devices.

The dataset and our focus on human optical flow opens up a number of research directions in human motion understanding and optical flow computation.
We would like to extend our dataset by modeling more diverse clothing and outdoor scenarios.
A direction of potentially high impact for this work is to integrate it in end-to-end systems for action recognition,
which typically take precomputed optical flow as input.
The real-time nature of the method could support motion-based interfaces, potentially even on devices like cell phones with limited computing power.
The data\neww{set, dataset generation code, pretrained} models, and training code are available, enabling researchers to \neww{use them for} problems involving human motion.

%% file: main.bbl
\begin{thebibliography}{10}

\bibitem{baker2011database}
Simon Baker, Daniel Scharstein, JP~Lewis, Stefan Roth, Michael~J Black, and
  Richard Szeliski.
\newblock A database and evaluation methodology for optical flow.
\newblock {\em International Journal of Computer Vision}, 92(1):1--31, 2011.

\bibitem{ranjan2016optical}
Anurag Ranjan and Michael~J Black.
\newblock Optical flow estimation using a spatial pyramid network.
\newblock In {\em Proc. of the IEEE Conference on Computer Vision and Pattern
  Recognition (CVPR)}, 2017.

\bibitem{Sun2018PWC-Net}
Deqing Sun, Xiaodong Yang, Ming-Yu Liu, and Jan Kautz.
\newblock {PWC-Net}: {CNNs} for optical flow using pyramid, warping, and cost
  volume.
\newblock In {\em CVPR}, 2018.

\bibitem{scienceselfies}
Donald Geman and Stuart Geman.
\newblock Opinion: Science in the age of selfies.
\newblock {\em Proceedings of the National Academy of Sciences},
  113(34):9384--9387, 2016.

\bibitem{Jhuang:ICCV:2013}
Hueihan Jhuang, Juergen Gall, Silvia Zuffi, Cordelia Schmid, and Michael~J.
  Black.
\newblock Towards understanding action recognition.
\newblock In {\em IEEE International Conference on Computer Vision (ICCV)},
  pages 3192--3199, Sydney, Australia, December 2013. IEEE.

\bibitem{soomro2012ucf101}
Khurram Soomro, Amir~Roshan Zamir, and Mubarak Shah.
\newblock Ucf101: A dataset of 101 human actions classes from videos in the
  wild.
\newblock {\em arXiv preprint arXiv:1212.0402}, 2012.

\bibitem{kuehne2011hmdb}
Hildegard Kuehne, Hueihan Jhuang, Est{\'\i}baliz Garrote, Tomaso Poggio, and
  Thomas Serre.
\newblock Hmdb: a large video database for human motion recognition.
\newblock In {\em Computer Vision (ICCV), 2011 IEEE International Conference
  on}, pages 2556--2563. IEEE, 2011.

\bibitem{Geiger2012CVPR}
Andreas Geiger, Philip Lenz, and Raquel Urtasun.
\newblock Are we ready for autonomous driving? the {KITTI} vision benchmark
  suite.
\newblock In {\em Conference on Computer Vision and Pattern Recognition
  (CVPR)}, 2012.

\bibitem{Butler:ECCV:2012}
D.~J. Butler, J.~Wulff, G.~B. Stanley, and M.~J. Black.
\newblock A naturalistic open source movie for optical flow evaluation.
\newblock In {A. Fitzgibbon et al. (Eds.)}, editor, {\em European Conf. on
  Computer Vision (ECCV)}, Part IV, LNCS 7577, pages 611--625. Springer-Verlag,
  October 2012.

\bibitem{dosovitskiy2015flownet}
Alexey Dosovitskiy, Philipp Fischery, Eddy Ilg, Caner Hazirbas, Vladimir
  Golkov, Patrick van~der Smagt, Daniel Cremers, Thomas Brox, et~al.
\newblock Flownet: Learning optical flow with convolutional networks.
\newblock In {\em 2015 IEEE International Conference on Computer Vision
  (ICCV)}, pages 2758--2766. IEEE, 2015.

\bibitem{ilg2016flownet}
Eddy Ilg, Nikolaus Mayer, Tonmoy Saikia, Margret Keuper, Alexey Dosovitskiy,
  and Thomas Brox.
\newblock Flownet 2.0: Evolution of optical flow estimation with deep networks.
\newblock {\em arXiv preprint arXiv:1612.01925}, 2016.

\bibitem{flyingthings}
N.Mayer, E.Ilg, P.H{\"a}usser, P.Fischer, D.Cremers, A.Dosovitskiy, and T.Brox.
\newblock A large dataset to train convolutional networks for disparity,
  optical flow, and scene flow estimation.
\newblock In {\em IEEE International Conference on Computer Vision and Pattern
  Recognition (CVPR)}, 2016.
\newblock arXiv:1512.02134.

\bibitem{Gaidon:Virtual:CVPR2016}
A~Gaidon, Q~Wang, Y~Cabon, and E~Vig.
\newblock Virtual worlds as proxy for multi-object tracking analysis.
\newblock In {\em CVPR}, 2016.

\bibitem{Bogo:ECCV:2016}
Federica Bogo, Angjoo Kanazawa, Christoph Lassner, Peter Gehler, Javier Romero,
  and Michael~J. Black.
\newblock Keep it {SMPL}: Automatic estimation of {3D} human pose and shape
  from a single image.
\newblock In {\em Computer Vision -- ECCV 2016}, Lecture Notes in Computer
  Science. Springer International Publishing, October 2016.

\bibitem{MANO:SIGGRAPHASIA:2017}
Javier Romero, Dimitrios Tzionas, and Michael~J. Black.
\newblock Embodied hands: Modeling and capturing hands and bodies together.
\newblock {\em ACM Transactions on Graphics, (Proc. SIGGRAPH Asia)}, 36(6),
  November 2017.

\bibitem{loper2014mosh}
Matthew Loper, Naureen Mahmood, and Michael~J Black.
\newblock Mosh: Motion and shape capture from sparse markers.
\newblock {\em ACM Transactions on Graphics (TOG)}, 33(6):220, 2014.

\bibitem{AMASS}
Naureen Mahmood, Nima Ghorbani, Nikolaus~F. Troje, Gerard Pons{-}Moll, and
  Michael~J. Black.
\newblock {AMASS:} archive of motion capture as surface shapes.
\newblock {\em CoRR}, abs/1904.03278, 2019.

\bibitem{Ranjan:BMVC:2018}
Anurag Ranjan, Javier Romero, and Michael~J. Black.
\newblock Learning human optical flow.
\newblock In {\em 29th British Machine Vision Conference}, September 2018.

\bibitem{Johansson1973}
Gunnar Johansson.
\newblock Visual perception of biological motion and a model for its analysis.
\newblock {\em Perception {\&} Psychophysics}, 14(2):201--211, 1973.

\bibitem{mhi_davis_2001}
J.~W. Davis.
\newblock Hierarchical motion history images for recognizing human motion.
\newblock In {\em Detection and Recognition of Events in Video}, pages 39--46,
  2001.

\bibitem{Black:IEEE:1997}
M.~J. Black, Y.~Yacoob, A.~D. Jepson, and D.~J. Fleet.
\newblock Learning parameterized models of image motion.
\newblock In {\em IEEE Conf. on Computer Vision and Pattern Recognition,
  CVPR-97}, pages 561--567, Puerto Rico, June 1997.

\bibitem{Black:ECCV:2002}
R.~Fablet and M.~J. Black.
\newblock Automatic detection and tracking of human motion with a view-based
  representation.
\newblock In {\em European Conf. on Computer Vision, ECCV 2002}, volume~1 of
  {\em LNCS 2353}, pages 476--491. Springer-Verlag, 2002.

\bibitem{Fragkiadaki2013}
K.~Fragkiadaki, H.~Hu, and J.~Shi.
\newblock Pose from flow and flow from pose.
\newblock In {\em 2013 IEEE Conference on Computer Vision and Pattern
  Recognition}, pages 2059--2066, June 2013.

\bibitem{Zuffi:ICCV:2013}
Silvia Zuffi, Javier Romero, Cordelia Schmid, and Michael~J Black.
\newblock Estimating human pose with flowing puppets.
\newblock In {\em IEEE International Conference on Computer Vision (ICCV)},
  pages 3312--3319, 2013.

\bibitem{PfisterCZ15}
Tomas Pfister, James Charles, and Andrew Zisserman.
\newblock Flowing convnets for human pose estimation in videos.
\newblock In {\em ICCV}, pages 1913--1921. IEEE Computer Society, 2015.

\bibitem{CharlesPMHZ16}
James Charles, Tomas Pfister, Derek~R. Magee, David~C. Hogg, and Andrew
  Zisserman.
\newblock Personalizing human video pose estimation.
\newblock In {\em CVPR}, pages 3063--3072. IEEE Computer Society, 2016.

\bibitem{flowcap}
Javier Romero, Matthew Loper, and Michael~J. Black.
\newblock {FlowCap}: {2D} human pose from optical flow.
\newblock In {\em Pattern Recognition, Proc. 37th German Conference on Pattern
  Recognition (GCPR)}, volume LNCS 9358, pages 412--423. Springer, 2015.

\bibitem{FeichtenhoferPZ16}
Christoph Feichtenhofer, Axel Pinz, and Andrew Zisserman.
\newblock Convolutional two-stream network fusion for video action recognition.
\newblock In {\em CVPR}, pages 1933--1941. IEEE Computer Society, 2016.

\bibitem{Dong_2018_CVPR}
Xuanyi Dong, Shoou-I Yu, Xinshuo Weng, Shih-En Wei, Yi~Yang, and Yaser Sheikh.
\newblock Supervision-by-registration: An unsupervised approach to improve the
  precision of facial landmark detectors.
\newblock In {\em The IEEE Conference on Computer Vision and Pattern
  Recognition (CVPR)}, 2018.

\bibitem{Freeman2000}
William~T. Freeman, Egon~C. Pasztor, and Owen~T. Carmichael.
\newblock Learning low-level vision.
\newblock {\em International Journal of Computer Vision}, 40(1):25--47, 2000.

\bibitem{wulff2015efficient}
Jonas Wulff and Michael~J Black.
\newblock Efficient sparse-to-dense optical flow estimation using a learned
  basis and layers.
\newblock In {\em 2015 IEEE Conference on Computer Vision and Pattern
  Recognition (CVPR)}, pages 120--130. IEEE, 2015.

\bibitem{roth2008learning}
Sun D., S~Roth, JP~Lewis, and MJ~Black.
\newblock Learning optical flow.
\newblock In {\em ECCV}, pages 83--97, 2008.

\bibitem{guney2016ACCV}
Fatma G{\"u}ney and Andreas Geiger.
\newblock Deep discrete flow.
\newblock In {\em Asian Conference on Computer Vision (ACCV)}, 2016.

\bibitem{epicflow}
Jerome Revaud, Philippe Weinzaepfel, Zaid Harchaoui, and Cordelia Schmid.
\newblock {EpicFlow: Edge-Preserving Interpolation of Correspondences for
  Optical Flow}.
\newblock In {\em {Computer Vision and Pattern Recognition}}, 2015.

\bibitem{Tran:End2End:2016}
Du~Tran, Lubomir Bourdev, Rob Fergus, Lorenzo Torresani, and Manohar Paluri.
\newblock Deep {End2End} {Voxel2Voxel} prediction.
\newblock In {\em The 3rd Workshop on Deep Learning in Computer Vision}, 2016.

\bibitem{shugrina2019creative}
Maria Shugrina, Ziheng Liang, Amlan Kar, Jiaman Li, Angad Singh, Karan Singh,
  and Sanja Fidler.
\newblock Creative flow+ dataset.
\newblock In {\em The IEEE Conference on Computer Vision and Pattern
  Recognition (CVPR)}, June 2019.

\bibitem{Gaidon2014}
Adrien Gaidon, Zaid Harchaoui, and Cordelia Schmid.
\newblock Activity representation with motion hierarchies.
\newblock {\em International Journal of Computer Vision}, 107(3):219--238,
  2014.

\bibitem{barbosa2018looking}
Igor~Barros Barbosa, Marco Cristani, Barbara Caputo, Aleksander Rognhaugen, and
  Theoharis Theoharis.
\newblock Looking beyond appearances: Synthetic training data for deep cnns in
  re-identification.
\newblock {\em Computer Vision and Image Understanding}, 167:50--62, 2018.

\bibitem{ghezelghieh2016learning}
Mona~Fathollahi Ghezelghieh, Rangachar Kasturi, and Sudeep Sarkar.
\newblock Learning camera viewpoint using cnn to improve 3d body pose
  estimation.
\newblock In {\em 2016 Fourth International Conference on 3D Vision (3DV)},
  pages 685--693. IEEE, 2016.

\bibitem{makehuman}
Makehuman: Open source tool for making 3d characters.

\bibitem{SMPL:2015}
Matthew Loper, Naureen Mahmood, Javier Romero, Gerard Pons-Moll, and Michael~J.
  Black.
\newblock {SMPL}: A skinned multi-person linear model.
\newblock {\em ACM Trans. Graphics (Proc. SIGGRAPH Asia)}, 34(6):248:1--248:16,
  October 2015.

\bibitem{surreal}
G\"{u}l Varol, Javier Romero, Xavier Martin, Naureen Mahmood, Michael Black,
  Ivan Laptev, and Cordelia Schmid.
\newblock Learning from synthetic humans.
\newblock In {\em CVPR}, 2017.

\bibitem{h36m_pami}
Catalin Ionescu, Dragos Papava, Vlad Olaru, and Cristian Sminchisescu.
\newblock Human3.6m: Large scale datasets and predictive methods for 3d human
  sensing in natural environments.
\newblock {\em IEEE Transactions on Pattern Analysis and Machine Intelligence},
  36(7):1325--1339, jul 2014.

\bibitem{cmumocapdatabase}
Carnegie-mellon mocap database.

\bibitem{Sigal:IJCV:10b}
L.~Sigal, A.~Balan, and M.~J. Black.
\newblock {HumanEva}: Synchronized video and motion capture dataset and
  baseline algorithm for evaluation of articulated human motion.
\newblock {\em International Journal of Computer Vision}, 87(1):4--27, March
  2010.

\bibitem{robinette2002civilian}
Kathleen~M Robinette, Sherri Blackwell, Hein Daanen, Mark Boehmer, and Scott
  Fleming.
\newblock Civilian american and european surface anthropometry resource
  (caesar), final report. volume 1. summary.
\newblock Technical report, DTIC Document, 2002.

\bibitem{cmumocap}
Ralph Gross and Jianbo Shi.
\newblock The cmu motion of body (mobo) database.
\newblock 2001.

\bibitem{yu_lsun}
Fisher Yu, Yinda Zhang, Shuran Song, Ari Seff, and Jianxiong Xiao.
\newblock Lsun: Construction of a large-scale image dataset using deep learning
  with humans in the loop.
\newblock {\em arXiv:1506.03365}, 2015.

\bibitem{xiao2010sun}
Jianxiong Xiao, James Hays, Krista~A Ehinger, Aude Oliva, and Antonio Torralba.
\newblock Sun database: Large-scale scene recognition from abbey to zoo.
\newblock In {\em Computer vision and pattern recognition (CVPR), 2010 IEEE
  conference on}, pages 3485--3492. IEEE, 2010.

\bibitem{spherical_harmonics}
Robin Green.
\newblock {Spherical Harmonic Lighting: The Gritty Details}.
\newblock {\em Archives of the Game Developers Conference}, March 2003.

\bibitem{collisionDeformableObjects}
Matthias Teschner, Stefan Kimmerle, Bruno Heidelberger, Gabriel Zachmann, Laks
  Raghupathi, Arnulph Fuhrmann, Marie-Paule Cani, Fran{\c{c}}ois Faure, Nadia
  Magnenat-Thalmann, Wolfgang Strasser, and Pascal Volino.
\newblock Collision detection for deformable objects.
\newblock In {\em Eurographics}, pages 119--139, 2004.

\bibitem{Dyna:SIGGRAPH:2015}
Gerard Pons-Moll, Javier Romero, Naureen Mahmood, and Michael~J. Black.
\newblock Dyna: A model of dynamic human shape in motion.
\newblock {\em ACM Transactions on Graphics, (Proc. SIGGRAPH)},
  34(4):120:1--120:14, August 2015.

\bibitem{Tzionas:IJCV:2016}
Dimitrios Tzionas, Luca Ballan, Abhilash Srikantha, Pablo Aponte, Marc
  Pollefeys, and Juergen Gall.
\newblock Capturing hands in action using discriminative salient points and
  physics simulation.
\newblock {\em International Journal of Computer Vision (IJCV)},
  118(2):172--193, June 2016.

\bibitem{he2017mask}
Kaiming He, Georgia Gkioxari, Piotr Doll{\'a}r, and Ross Girshick.
\newblock Mask r-cnn.
\newblock In {\em Computer Vision (ICCV), 2017 IEEE International Conference
  on}, pages 2980--2988. IEEE, 2017.

\bibitem{matterport_maskrcnn_2017}
Waleed Abdulla.
\newblock Mask r-cnn for object detection and instance segmentation on keras
  and tensorflow.
\newblock \url{https://github.com/matterport/Mask_RCNN}, 2017.

\bibitem{kingma2014adam}
Diederik Kingma and Jimmy Ba.
\newblock Adam: A method for stochastic optimization.
\newblock {\em arXiv preprint arXiv:1412.6980}, 2014.

\bibitem{he2015deep}
Kaiming He, Xiangyu Zhang, Shaoqing Ren, and Jian Sun.
\newblock Deep residual learning for image recognition.
\newblock {\em arXiv preprint arXiv:1512.03385}, 2015.

\bibitem{menze2015object}
Moritz Menze and Andreas Geiger.
\newblock Object scene flow for autonomous vehicles.
\newblock In {\em Proceedings of the IEEE Conference on Computer Vision and
  Pattern Recognition}, pages 3061--3070, 2015.

\bibitem{brox2009large}
Thomas Brox, Christoph Bregler, and Jitendra Malik.
\newblock Large displacement optical flow.
\newblock In {\em Computer Vision and Pattern Recognition, 2009. CVPR 2009.
  IEEE Conference on}, pages 41--48. IEEE, 2009.

\bibitem{flowfields}
Christian Bailer, Bertram Taetz, and Didier Stricker.
\newblock Flow fields: Dense correspondence fields for highly accurate large
  displacement optical flow estimation.
\newblock In {\em Proceedings of the IEEE International Conference on Computer
  Vision}, pages 4015--4023, 2015.

\bibitem{Felzenszwalb2010PAMI}
P.~F. Felzenszwalb, R.~B. Girshick, D.~McAllester, and D.~Ramanan.
\newblock Object detection with discriminatively trained part-based models.
\newblock {\em TPAMI}, 32(9):1627--1645, 2010.

\bibitem{MilanL0RS16}
Anton Milan, Laura Leal{-}Taix{\'{e}}, Ian~D. Reid, Stefan Roth, and Konrad
  Schindler.
\newblock {MOT16:} {A} benchmark for multi-object tracking.
\newblock {\em arXiv:1603.00831}, 2016.

\end{thebibliography}
